\documentclass[10pt,twocolumn,letterpaper]{article}

\usepackage[pagenumbers]{cvpr} %

\usepackage{times}
\usepackage{epsfig}
\usepackage{graphicx}
\usepackage{amsmath}
\usepackage{amssymb}
\usepackage{multirow}
\usepackage{tabularx}
\usepackage{makecell}
\usepackage{booktabs}
\usepackage{mathtools}
\usepackage{pgfplots}

\usepackage{subcaption}
\usepackage{algpseudocode,algorithm,algorithmicx}
\usepackage{xr}
\usepackage{bm}
\usepackage{color}
\usepackage[dvipsnames]{xcolor}
\usepackage{nicefrac}       %
\usepackage{microtype}      %

\usepackage{enumitem}
\usepackage{csquotes}
\usepackage{siunitx}
\usepackage{comment}

\newcommand{\lstart}{\mathrm{start}} %
\newcommand{\lend}{\mathrm{end}} %

\newcommand{\polarity}{p} %
\newcommand{\probability}{\mathbb{P}} %
\newcommand{\posi}[1]{{{#1}^+}} %
\newcommand{\nega}[1]{{{#1}^-}} %
\newcommand{\pprobpos}{\posi{q}} %
\newcommand{\pprobneg}{\nega{q}} %
\newcommand{\accpol}{E} %
\newcommand{\numevents}{n} %
\newcommand{\expectation}{\mathbb{E}} %
\newcommand{\variance}{\mathbb{V}} %
\newcommand{\jeq}[1]{\stackrel{#1}{=}} %

\newcommand{\reals}{\mathbb{R}} %
\newcommand{\decay}{b} %

\newcommand{\contract}{\hspace{-0.5cm}} %

\renewcommand{\paragraph}[1]{\noindent\textbf{#1.}}

\definecolor{cvprblue}{rgb}{0.21,0.49,0.74}
\usepackage[pagebackref,breaklinks,colorlinks,allcolors=cvprblue]{hyperref}
\usepackage[capitalize]{cleveref}
\Crefname{equation}{Eq.}{Eqs.}
\creflabelformat{equation}{#2#1#3}
\Crefname{figure}{Fig.}{Figs.}
\Crefname{tabular}{Tab.}{Tabs.}
\Crefname{section}{Sec.}{Secs.}
\Crefname{appendix}{Appx.}{Appx.}

\def\paperID{4} %
\def\confName{CVPR}
\def\confYear{2025}

\title{Dynamic EventNeRF: Reconstructing General Dynamic Scenes \\ from Multi-view RGB and Event Streams} 

\author{
    Viktor Rudnev$^{1,2}$ \hspace{0.2em}
    Gereon Fox$^{1,2}$  \hspace{0.2em}
Mohamed Elgharib$^1$ \hspace{0.2em}
Christian Theobalt$^1$ \hspace{0.2em}
Vladislav Golyanik$^1$ \vspace{7pt} \\
\hspace{-50pt}
$^1$MPI for Informatics, SIC \hspace{40pt}
$^2$Saarland University
}

\begin{document}

\makeatletter
\g@addto@macro\@maketitle{
  \begin{figure}[H]
  \vspace{-15pt}
  \setlength{\linewidth}{\textwidth}
  \setlength{\hsize}{\textwidth}
  \centering
\includegraphics[width=0.95\textwidth,trim={0cm 0cm 0cm 0cm},clip]{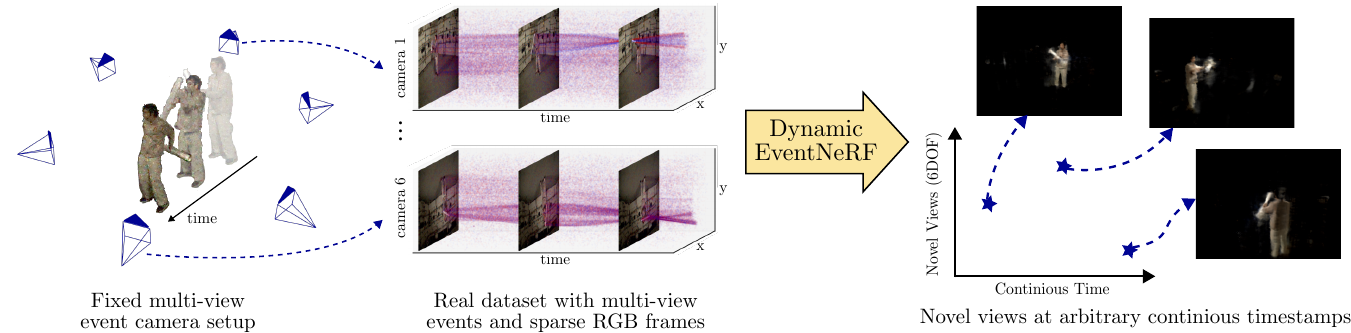}
\caption{\textbf{Dynamic EventNeRF} is the first approach to reconstruct general dynamic scenes in 4D using \textbf{multi}-view event streams and sparse RGB frames. Our method produces novel views at arbitrary timestamps of $\ang{360}$ scenes with fast motion and challenging lighting conditions.
    }
    \label{fig:teaser}
  \end{figure}
}
\makeatother

\maketitle
\begin{abstract}
Volumetric reconstruction of dynamic scenes is an important problem in computer vision.
It is especially challenging in poor lighting and with fast motion.
This is partly due to the limitations of RGB cameras:
To capture frames under low lighting, the exposure time needs to be increased, which leads to more motion blur.
In contrast, event cameras, which record changes in pixel brightness asynchronously, are much less dependent on lighting, making them more suitable for recording fast motion.
We hence propose the first method to spatiotemporally reconstruct a scene from sparse multi-view event streams and sparse RGB frames. 
We train a sequence of cross-faded time-conditioned NeRF models, one per short recording segment.
The individual segments are supervised with a set of event- and RGB-based losses and sparse-view regularisation.
We assemble a real-world multi-view camera rig with six static event cameras around the object and record a benchmark multi-view event stream dataset of challenging motions. 
Our work outperforms RGB-based baselines, producing state-of-the-art results, and opens up the topic of multi-view event-based reconstruction as a new path for fast scene capture beyond RGB cameras.
\end{abstract}

\vspace{-10pt}
\section{Introduction}

\label{sec:intro}
\emph{Spatiotemporal (or 4D) reconstruction} of a non-rigid scene allows re-rendering it from novel viewpoints.
This is a long-standing and challenging problem in computer vision \cite{yunus2024nonrigidstar,Tung2009CompleteMR}.
Until recently, the great majority of methods used RGB frames as input data. However, RGB frames contain motion blur for fast objects and become noisier the less light is available in the scene.
This has motivated the exploration of event cameras \cite{zheng2023events,chakravarthi2024events,gallego2022events} as a data source for this task, which have greater dynamic range and superior temporal resolution.
Since event cameras are by far not as commonplace as RGB cameras and thus still more expensive, previous event-based non-rigid spatiotemporal capture techniques~\cite{ma2023deformable,rudnev2021eventhands,zou2021eventhpe,xue2022handsbmvc,EventCap2020,Millerdurai_EventEgo3D_2024} are mostly monocular.
Using only one camera can be expected to limit the ability to capture general dynamic scenes, especially when handling fast motion, large deformations and occlusions~\cite{gao2022realitycheck}.

In this work, we explore the use of \textbf{multi}-view event data for event-based 4D reconstruction of non-rigid motion.
Since a multi-view event camera setup is, of course, more complex than a monocular setup, we believe that investigating this uncharted territory can prove a helpful guideline in deciding whether the added cost is worth the gains in quality.

While previous work \cite{rudnev2023eventnerf,hwang2023evnerf,klenk2023enerf} addressed the event-based reconstruction of \emph{static} scenes, 
this work reconstructs \emph{dynamic} scenes, \ie{} yields different 3D models for different time points.
A common approach in reconstructing dynamic scenes is to explain each state as a deformation of a canonical volume~\cite{park2021nerfies,park2021hypernerf,tretschk2021nonrigid}.
However, this tends to limit both the range and kinds of reconstructed motion~\cite{TretschkNonRigidSurvey,gao2022realitycheck}.
To address these challenges,
we train a sequence of cross-faded time-conditioned NeRF models, each representing a short recording segment.
This way, smaller motions can be learned locally with consistent quality, even if the full recording is long.
Moreover, having no explicit deformation models, we can reconstruct general motion.

To evaluate our method, we both render synthetic data and obtain real data by setting up a multi-view rig of six event cameras, with which we record a benchmark dataset of various scenes, subjects and motions under lighting conditions ranging from \enquote{dimly lit} to \enquote{very dark}~(\cref{fig:teaser}).
We compare our method to RGB-based baselines trained on blurry RGB videos and RGB videos reconstructed from events.
Our method significantly outperforms all baselines and produces state-of-the-art results, while providing a continuous reconstruction of the scene.
It demonstrates that operating directly on events instead of reconstructed RGB images in challenging conditions results in higher novel-view rendering accuracy.
To summarise, our contributions are as follows:
\begin{itemize}[leftmargin=15pt]
 \item[1)] We present Dynamic EventNeRF, the first method for learning general 4D scenes from sparse \textbf{multi}-view event streams and sparse blurry RGB frames;
 \item[2)] We demonstrate the usage of time-conditioned NeRF models, our blended multi-segment approach, an event-based supervision scheme, and fast event accumulation with decay to learn the sequences;
 \item[3)] We provide multi-view static camera real and synthetic datasets for 4D reconstruction from event streams. %
\end{itemize}
Our datasets and the source code are available online\footnote{\href{https://4dqv.mpi-inf.mpg.de/DynEventNeRF/}{https://4dqv.mpi-inf.mpg.de/DynEventNeRF/}}. 

\section{Related Works}

\paragraph{NeRF for non-rigid scenes with RGB inputs and sparse views}
Novel view synthesis has seen recent progress \cite{Mildenhall19,sitzmann2019srns,Lombardi:2019,mildenhall2020nerf,kerbl20233dgs}, with methods based on Neural Radiance Fields (NeRF)~\cite{mildenhall2020nerf}  being especially successful.
NeRF has been adapted for dynamic scene reconstruction, \eg~head avatars~\cite{prao2022vorf}, full-body avatars~\cite{peng2021neuralbody,liu2021neural,hdhumans}, hands~\cite{mundra2023livehand} and other domains.
Other techniques target general scene reconstruction, which can be divided into monocular~\cite{pumarola2021dnerf,park2021hypernerf,tretschk2021nonrigid,li2023dynibar} and multi-view methods~\cite{fridovich2023kplanes,cao2023hexplane,shao2023tensor4d,peng2023mlpmaps,neus2,tretschk2023scenerflow}.
While the monocular methods can reconstruct the input video with high fidelity, the ambiguities of monocular input make it hard for them to learn the correct geometry, greatly limiting how far novel views can deviate from the training viewpoints.
Moreover, many monocular methods greatly depend on the camera moving faster than the object, constraining the scenes they can reconstruct~\cite{gao2022realitycheck}.
As we capture fast motion, moving the camera even faster is not a viable option.
Our method uses a fixed camera multi-view setup.
Multi-view methods generalise better to novel views, but are still limited by their RGB input, which is prone to motion blur and noise, especially with fast motion or low lighting conditions.
Instead, we use event cameras, possessing both high temporal resolution and high dynamic range.
To train NeRF models, one typically needs dozens of ground-truth RGB views, but in our setting, we only have six views available.
Follow-up works of NeRF aim at enabling few-view reconstruction, using extensive regularisation~\cite{niemeyer2022regnerf}, pre-training~\cite{yu2020pixelnerf,mueller2022autorf}, semantic consistency~\cite{dietnerf}, or depth priors~\cite{neff2021donerf,roessle2022depthpriorsnerf}.
Our approach (\cref{ssec:supervision}) uses a combination of regularisers and volume clipping, without a need for pretraining or additional priors, and is applicable to general scenes.

\paragraph{Event-based vision for dynamic scenes and 3D reconstruction}
Event cameras have been used to reconstruct non-blurry RGB videos of fast motion: Some methods \cite{Rebecq19pami,Paredes21} use a learning-based approach to convert events into RGB frames, while others \cite{pan2019bringing,Tulyakov21CVPR,tulyakov2022timelenspp} combine events and input RGB frames and use the former to deblur and interpolate the latter.
However, all of them only support 2D reconstruction.
Methods based on 3D meshes and parametric models allow tracking of full bodies~\cite{EventCap2020,zou2021eventhpe,dhp19dataset}, hands~\cite{rudnev2021eventhands,Nehvi2021,xue2022handsbmvc,Jiang2024EvRGBHand,Millerdurai_3DV2024} and general templates~\cite{Nehvi2021,xue2022handsbmvc}, but do not reconstruct appearance and are not applicable to \emph{general} scenes, which is what our \emph{Dynamic EventNeRF} is aiming for.

For general static scenes, the state of the art is represented by a new class of NeRF-based methods that allow dense 3D reconstruction and novel-view synthesis using event streams:
EventNeRF~\cite{rudnev2023eventnerf} and Ev-NeRF~\cite{hwang2023evnerf} supervise NeRF using accumulated windows of events, E-NeRF~\cite{klenk2023enerf} supervises each event directly, Robust E-NeRF\cite{low2023robustenerf} extends it to account for the pixel refraction and noise, E\textsuperscript{2}NeRF~\cite{qi2023e2nerf} and Ev-DeblurNeRF~\cite{Cannici_2024_CVPR} use events to enhance and deblur RGB-based scene reconstructions.
These methods model static scenes using data captured with a single moving camera, while our Dynamic EventNeRF reconstructs \emph{general dynamic} scenes using a \emph{multi-view fixed camera} setup.

DE-NeRF~\cite{ma2023deformable} reconstructs dynamic scenes with monocular event streams and RGB frames from a moving camera through deformations of a canonical volume.
As discussed earlier, such a setting and approach significantly limit the kind of scenes and motions that can be reconstructed.
In contrast, our Dynamic EventNeRF uses  a \emph{multi-view fixed camera setup} and is \emph{not based on canonical volumes}, allowing it to reconstruct \emph{general dynamic scenes}.

EvDNeRF~\cite{bhattacharya2024evdnerf} allows to synthesise events from novel viewpoints using multi-view event streams through the deformation of a canonical volume.
It does not model the appearance, needs as many as 18 views, and the deformation-based approach significantly limits the scenes and motions that can be reconstructed.
In contrast, our Dynamic EventNeRF allows to reconstruct \emph{general dynamic scenes directly in RGB space} while using \emph{as few as three views}.

\section{Background}
\subsection{Event Camera Model and Deblurring}
\label{ssec:evcmodel}

Instead of capturing dense RGB frames at regular intervals, event cameras operate asynchronously per individual pixel:
When a certain pixel gets brighter or darker by a certain threshold, it emits an \textit{event}, which is a tuple $(t, x, y, p)$ of its timestamp $t$, pixel position $x, y$, and polarity $p \in \{-1, +1\}$. The event can be interpreted as the assertion
\begin{equation}
    \log I_{x,y}(t)-\log I_{x, y}(t')=pC_{p},
\label{eq:eventgeneration}
\end{equation}
where $\log I_{x,y}(\tau)$ is the intensity of pixel $(x, y)$ at time $\tau$ and $t'<t$ is the timestamp of the previous event in that pixel. We call  $C_\mathrm{+1}, C_\mathrm{-1}>0$ the event generation thresholds.
\emph{Event-based Single Integral} (ESI)~\cite{rudnev2023eventnerf} connects the difference between logarithmic intensities on its left side and the accumulated events on the right side by summing \cref{eq:eventgeneration} over time.
ESI can be further extended to Event-based Double Integral (EDI)~\cite{pan2019bringing} by averaging the intensity over the exposure period, to model motion blur caused by the traditional camera shutter.
Given events $\mathbf{E}$ and the blurry image $\mathbf{B}$, one can solve a system of equations generated by EDI for the unknown instantaneous intensity image $\mathbf{I}$, thus deblurring $\mathbf{B}$.
This way, EDI can be used for deblurring the RGB stream through events.
To synthesise arbitrary instantaneous frames $\mathbf{I}(t)$, one could accumulate events on top of the closest recovered deblurred frame $\mathbf{I}(t_0)$ with ESI.
To improve the speed of deblurring and frame synthesis, Lin et al.~propose Fast EDI~\cite{fastedi}, which uses the same event camera model and a special data structure to accelerate the queries.

\subsection{EventNeRF}
\label{ssec:eventnerf}
Rudnev et al.~\cite{rudnev2023eventnerf} represent a static scene as an MLP predicting point density $\sigma(\mathbf{p}) \in \mathbb{R}$ and colour $\mathbf{C}(\mathbf{p}, \mathbf{\theta})$ as functions of its position $\mathbf{p} \in \mathbb{R}^3$ and view directions $\theta \in \mathbb{R}^2$.
Novel views are obtained through volumetric rendering.
First, they associate a ray $\mathbf{r}^{(x,y,t)}(d)\in \mathbb{R}^3$ with every pixel $(x,y)$ that they render from the viewpoint at time $t$.
Then they randomly sample them at depths $\{d_i\}_{i=1}^{N_\textrm{samples}}$: $\{\mathbf{r}^{(x,y,t)}_i\}_{i=1}^{N_\textrm{samples}}$.
And finally, they integrate them into the pixel colour as:
\begin{equation}
    \mathbf{\hat C}_{x,y,t}=\sum_{i=1}^{N_\textrm{samples}} (1-\exp(-\sigma_i))T_i \mathbf{C}_i,
    \label{eq:nerf}
\end{equation}
where $T_i=\exp\left(-\sum_{j=1}^{N_\textrm{depth}-1}\sigma_j(d_{j+1}-d_j)\right)$ is the accumulated transmittance, $d_j$ is the distance of the $j^\textrm{th}$ ray sample from the ray origin, $\sigma_i=\sigma\left(\mathbf{r}_i^{(x,y,t)}\right)$ and $\mathbf{C}_{i}=\mathbf{C}\left(\mathbf{r}^{(x,y,t)}_i, \mathbf{\theta}^{(x,y,t)}\right)$ are the sample density and colour.

To supervise this model with ground-truth events, the authors use ESI~(\cref{ssec:evcmodel}):
They substitute intensity values $I_{x,y}(t)$ with the rendered predictions $\hat{\mathbf{C}}_{x,y,t}$ and apply the MSE loss between the intensity and event sides of ESI:

\begin{equation}
    \mathcal L_{x,y}(t_0,t)={\|\hat{E}_{x,y}(t_0, t)-E_{x,y}(t_0, t)\|^2},
\label{eq:lrender}
\end{equation}
where $\hat{E}_{x,y}(t_0, t)=\mathbf{F}(\log \hat{\mathbf{C}}_{x,y,t}-\log \hat{\mathbf{C}}_{x,y,t_0})$ is the predicted difference, $E(t_0, t)=\sum_{i: t_0 < t_i \leq t} {p_iC_{p_i}\delta_{x_i,y_i}^{x,y}}$ is the accumulated difference, $\mathbf{F}$ is the colour filter, $\delta_{\mathbf{a}}^{\mathbf{\hat{a}}}=1$ iff. $\mathbf{a}=\mathbf{\hat{a}}$, $0$ otherwise.
The performance is also much improved with \emph{Positional Encoding} \cite{mildenhall2020nerf}.

\section{Our Dynamic EventNeRF Approach}
\begin{figure*}[t]
\centering
\includegraphics[width=0.95\textwidth,trim={0cm 0cm 0cm 0cm},clip]{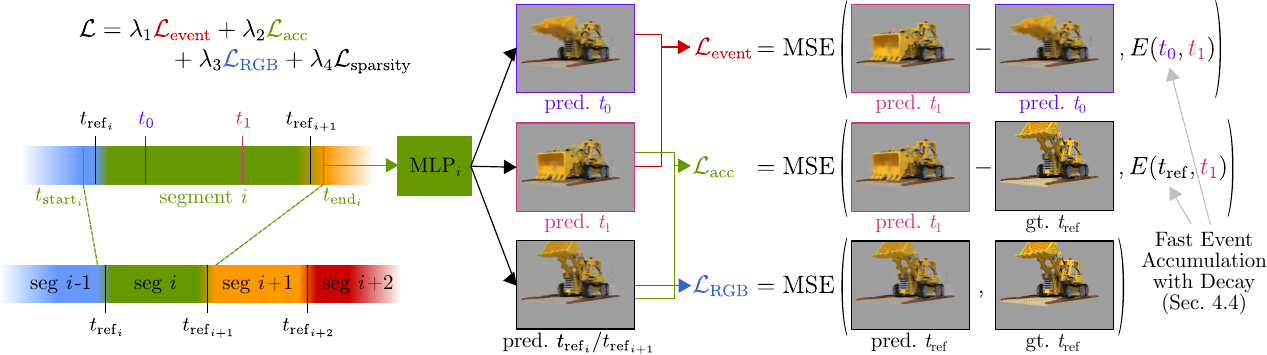}
\caption{Overview of \textbf{Dynamic EventNeRF}. We split the entire sequence into short overlapping segments (\eg, $\mathrm{seg}~i-1$, $\mathrm{seg}~i$, $\mathrm{seg}~i+1$, $\mathrm{seg}~i+2$ on the bottom-left of the figure). For each segment, we learn a time-conditioned MLP-based NeRF model. To supervise it, we first sample a random window $[t_0, t_1]$ within the segment and apply a combination of the following losses: 1) \emph{Event loss}, supervising predicted view differences; 2) \emph{Accumulation loss}, supervising differences between a reference RGB frame and one of the predicted views; 3) \emph{RGB loss}, supervising the model with the reference RGB frame, and 4) \emph{Sparsity loss}, that minimises the number of opaque pixels in each predicted views. For increased stability and fast computation of these losses, we propose \emph{Fast Event Accumulation with Decay}; see \cref{ssec:accumulation}.
    }
    \label{fig:method}
\end{figure*}

We are interested in novel-view synthesis based on sparse-view event streams: We have a sparse set of $K$ event cameras recording a general dynamic scene from $T=0$ to time $T_\mathrm{end}$.
As we target full scene reconstruction, no part of it should be left unobserved, and thus the cameras cover $360^\circ$ of the subject.
Since the appearance of the background influences the foreground event polarity, we capture a set of RGB images $\{\mathbf{A}^k\}_{k=1}^K$ of the background without the subject.
Moreover, we use supporting RGB frames $\{\mathbf{C}^k(t_{\mathrm{ref}_j})\}_{k=1,0\leq t_{\mathrm{ref}_j}\leq T_\mathrm{end}}^K$ captured at regular but sparse intervals of time, e.g., $\qty{1}{\s}$, and then deblurred through Fast EDI~\cite{fastedi}.
To reconstruct a 4D model of the sequence from the recorded data, we propose our Dynamic EventNeRF model (\cref{ssec:onesegmentmodel}, \cref{fig:method}).
The opacity and colour for each point in space-time are captured by a temporally-conditioned MLP-based NeRF.
As output, we render arbitrary novel views of the scene at arbitrary moments of time.
Since the capacity of an MLP model is limited,  we split longer sequences into smaller segments, and train a separate model for each segment (\cref{ssec:multisegmentmodel}).
To train the models, we propose an event-based sampling scheme and losses, and a set of regularisation techniques related to the sparse-view setup (\cref{ssec:supervision}).
Additionally, we propose an acceleration technique, enabling us to efficiently implement the proposed event supervision (\cref{ssec:accumulation}).

\subsection{Model for One Segment}
\label{ssec:onesegmentmodel}
We reconstruct a 4D model of the $i^\mathrm{th}$ temporal interval $[t_{\mathrm{start}_i}, t_{\mathrm{end}_i}]$, with $0\leq t_{\mathrm{start}_i} \leq t_{\mathrm{end}_i} \leq T_\mathrm{end}$,
using sparse multi-view events $\{\mathbf{E}^k\}_{k=1}^K$ and the supporting instantaneous (non-blurry) RGB views $\{\mathbf{C}^k(t_{\mathrm{ref}_{i}}), \mathbf{C}^k(t_{\mathrm{ref}_{i+1}})\}_{k=1}^K$ at the start and the end of the interval (\cref{fig:method}, left).
For convenience of notation, we linearly map the temporal extent of the interval to $-1 \leq t \leq 1$.
To represent the scene, we use an MLP mapping space-time coordinates $\mathbf{p}=[x, y, z, t]^T$ to density $\sigma \in \mathbb{R}$ and colour $\mathbf{C} \in \mathbb{R}^3$, where $-1 \leq x, y, z, t \leq 1$.
We obtain the foreground ray colours $\mathbf{\hat C}^{k,\textrm{fg}}$ of the training view $k$ through volumetric rendering of $\sigma$ and $\mathbf{C}$, similar to NeRF.
To obtain the final ray colours $\mathbf{\hat C}$, we alpha-blend the foreground and background:
\begin{equation}
    \mathbf{\hat C}^k_{x,y}=\mathbf{\hat C}_{x,y}^{k,\textrm{fg}}+T_{N_\mathrm{samples}+1}\cdot \mathbf{A}^k_{x,y},
\end{equation}
where $T_{N_\mathrm{samples}+1}$ represents background opacity (\cref{eq:nerf}).
Note that instead of learning the background with a separate model, we use the ground-truth background $\mathbf{A}^k$ for predictions of the view $k$.
This eliminates possible artefacts and ambiguities due to imprecise background modelling.%
 To facilitate the training, we use separate Positional Encodings for embedding time and spatial coordinates $\mathbf{p}$
(\cref{ssec:eventnerf}).
The maximum frequency of the encoding determines how coherent the model is across the corresponding dimension.
If it is zero, the model would have no variation over that dimension, and if it is too high, then the model would behave randomly due to aliasing.
We choose the optimal values empirically to be 7 temporal and 14 spatial frequencies.

Using a single MLP network to represent longer periods of time would pose a number of challenges:
The network would have to be sufficiently large, significantly slowing down both  inference and training.
Moreover, supervising the MLP at one moment of time $t$, could, as an unwanted side effect, cause artefacts at $t'\neq t$, as both times are represented with a single network.
In the next section, we overcome these issues by splitting the full sequence into smaller segments and training one model per segment.

\subsection{Multi-Segment Model}
\label{ssec:multisegmentmodel}

Using a single MLP to represent longer sequences is challenging due to the capacity-performance trade-off and the ghosting artefacts caused by using a shared network.
Therefore, we propose to split the full sequence into short segments and train one model per such sequence.
To eliminate flickering caused by changing from one model to the next, we overlap the segments by $10\%$ at the starts and at the ends.
In the overlapped parts, we render both models as $\mathbf{\hat C}_i(t)$ and $\mathbf{\hat C}_{i+1}(t)$ and cross-fade the predictions:
\begin{equation}
    \mathbf{\hat C}(t) = (1-\alpha(t)) \mathbf{\hat C}_i(t) + \alpha(t) \mathbf{\hat C}_{i+1}(t),
\end{equation}
where $\alpha(t)=\frac{t-t_\textrm{ov. start}}{t_\textrm{ov. end}-t_\textrm{ov. start}}$, with $t_\textrm{ov. start}$ and $t_\textrm{ov. end}$ being the overlapping start and end times, respectively.
In the next section, we describe the training procedure and all the regularisation techniques we use to adapt the method to event supervision and the sparse-view setting.

\subsection{Supervision for One Segment}
\label{ssec:supervision}
In this section, we describe how we supervise the model for a single segment $i$ from $t_{\mathrm{start}_i}$ to $t_{\mathrm{end}_i}$ (\cref{fig:method}).
We use sparse multi-view events $\{\mathbf{E}^k_i\}_{k=1}^K$ and the supporting instantaneous RGB views $\{\mathbf{C}^k(t_{\mathrm{ref}_{i}}), \mathbf{C}^k(t_{\mathrm{ref}_{i+1}})\}_{k=1}^K$ at the start and the end of the segment.

\paragraph{Event supervision}
For event input, we start by randomly choosing a temporal window $[t_0, t_1]$:
$t_1$ is sampled from the uniform distribution $U[t_{\mathrm{start}_i}, t_{\mathrm{end}_i}]$;
$t_0$ is chosen such that the window duration $t_1-t_0$ is distributed uniformly from $10\%$ to $30\%$ of the segment's length.
For each temporal window, per each view $1 \leq k \leq K$, we accumulate the events inside the window into $E^k(t_0, t_1) \in \mathbb{R}^{W\times H}$ according to the model described in \cref{ssec:accumulation}%

Similarly to EventNeRF~\cite{rudnev2023eventnerf}, we form training batches by combining $10\%$ positive and $90\%$ negative rays jointly for each view $k$:
Positive rays are randomly chosen pixels $x,y$ for which $E_{x,y}(t_0, t_1) \neq 0$, \ie from regions where events occurred.
Negative rays are the pixels uniformly sampled from all pixels in the view.
Including them allows us to reduce noise and reconstruct uniformly coloured details.
For the colour values obtained for the sampled rays,  we compute an MSE loss $L_\mathrm{event}$ in the event camera log-space: 
\begin{equation}
\small
\begin{split}
     & K \cdot \mathcal{L}_\mathrm{event} = \\
     &\sum_{k=1}^K \mathrm{MSE} \left(  E^k(t_0, t_1),  F\left(\log{\mathbf{\hat C}}^k(t_1)-\log{\mathbf{\hat C}}^k(t_0)\right)\right),
\end{split}
\label{eq:eventloss}
\end{equation}
where $\mathbf{\hat{C}}^k(t) \in \mathbb{R}^3$ is the model prediction at time $t$ from view $k$, and $F(\cdot): \mathbb{R}^{H\times W \times 3} \rightarrow \mathbb{R}^{H\times W}$ is applying the  Bayer filter to the RGB image, resulting in a greyscale intensity image matching the event camera colour filter.

\paragraph{RGB supervision}
To make the model match RGB ground-truth values $\{\mathbf{C}^k(t_{\mathrm{ref}})\}_{k=1}^K$, $t_\mathrm{ref} \in \{t_{\mathrm{ref}_i}, t_{\mathrm{ref}_{i+1}} \}$, we compute another MSE loss:
\begin{equation}
    \mathcal{L}_\mathrm{RGB} = \frac{1}{K}\sum_{k=1}^K \mathrm{MSE}\left(\mathbf{C}^k(t_\mathrm{ref}), {\mathbf{\hat C}}^k(t_\mathrm{ref})\right),
    \label{eq:rgbloss}
\end{equation}
    where ${\mathbf{\hat C}}^k(t_\mathrm{ref})$ is the model prediction.

\paragraph{Accumulation loss}
\cref{eq:rgbloss} only supervises the model at the reference RGB frame timestamps.
Because of that, the information from the reference frame is not propagated well to the rest of the sequence.
To mitigate this issue, we also use the \emph{accumulation loss}, which connects the closest reference RGB frame $\mathbf{C}^k(t_{\mathrm{ref}})$, the prediction at the end of the window $\mathbf{\hat C}^k(t_1)$, and the accumulated events $E^k(t_{\mathrm{ref}}, t_1)$ in between:
\begin{equation}
\small
\begin{split}
    & K \cdot \mathcal{L}_\mathrm{acc} = \\
    & \sum_{k=1}^K \mathrm{MSE}\left(E^k(t_\mathrm{ref}, t_1), F\left(\log{\mathbf{\hat C}}^k(t_1)-\log{\mathbf{C}}^k(t_\mathrm{ref})\right)\right).
\end{split}
\label{eq:accloss}
\end{equation}

\paragraph{Sparse-view regularisation}
For sparse-view reconstruction, we use the following combination of regularisation techniques:
First, we want to prevent the model from learning artefacts in the regions where only one view observes them.
These areas are near and far away from the camera, but still in its frustum.
Suppose the views are distributed $\ang{360}$ around the object at roughly similar height: The intersection of all frustums can be approximated by a cylinder of radius  $r$ that extends vertically from $y_\mathrm{min}$ to $y_\mathrm{max}$. We can then clip the reconstruction volume by setting $\sigma(x,y,z) := 0$ wherever $x^2+z^2>r$, or $y > y_\textrm{max}$ or $y < y_\textrm{min}$.
We also modify the ray depth sampling so that the samples are only chosen inside the cylinder.
Following EventNeRF, we randomly offset the ray direction within a pixel to improve generalisation to novel views.
This cylinder clipping and the improved sampling procedure result in a strong prior significantly reducing the ambiguity of the sparse-view reconstruction.

As stated earlier, we want our model only to reconstruct the foreground.
Since we use the cylindrical bounding volume, the model does not have a way to represent the background (that is out of bounds) correctly.
However, the model can still create background-coloured artefacts within the bounds, which it can also use to hide view-inconsistent artefacts.
To resolve these issues, we  minimise the number of rays hitting the foreground (or, equivalently, maximise the number of rays hitting the actual background) by applying a per-ray sparsity loss:
\begin{equation}
    \mathcal L_\mathrm{sparsity}(n)=\gamma_n\cdot\frac{1}{WH}\sum_{{x,y}=1}^{WH} \left(1-T_{N_\mathrm{samples}+1}^{(x,y)}\right),
\end{equation}
where $W,H$ are width and height of the image, $T_{N_\mathrm{samples}+1}$ represents background opacity (\cref{eq:nerf}), $\gamma_n=1-\exp\left(\frac{n}{N_\mathrm{sp. anneal}}\right)$, and $n$ is the training iteration and $N_\mathrm{sp. anneal}=\num{4d4}$.
Note that we gradually anneal the regularisation strength through the factor $\gamma_n$.
If we applied the regulariser at its full strength from the start, it might overpower the other loss terms before there is any meaningful reconstruction in place, resulting in the model converging to a blank scene.

Learning high-frequency details from the start is not optimal, especially in the sparse-view setting.
Thus, we implement coarse-to-fine training by applying positional encoding annealing~\cite{park2021nerfies} both across spatial frequencies and time.
Annealing improves temporal and spatial smoothness as the method starts with a coarse representation, slowly progressing towards a finer representation.

\paragraph{Full loss}
We combine all aforementioned losses:
\begin{equation}
    \mathcal{L}=\lambda_1\mathcal{L}_\mathrm{event}+
    \lambda_2\mathcal{L}_\mathrm{acc}+\lambda_3\mathcal{L}_\mathrm{RGB}+\lambda_4\mathcal{L}_\mathrm{sparsity},
\end{equation}
where $\lambda_1=\num{1}$, $\lambda_2=\num{e-2}$, $\lambda_3=\num{1}$, and $\lambda_4=\num{e-2}$.
In the next section, we describe how we compute the accumulated events $E(t_0, t_1)$, propose how to modify accumulation to work with longer windows without diverging, and how to accelerate the computation so that we can quickly evaluate $E(\cdot, \cdot)$ every training iteration.

\subsection{Event Accumulation}
\label{ssec:accumulation}

Following \cref{ssec:evcmodel}, we  define event accumulation:
\begin{equation}\label{eq:accumulation}
    E_{x,y}(t_0, t_1) = \sum_{i=i_\lstart}^{i_\lend}  \delta_{x_i,y_i}^{x,y} p_i C_{p_i} \decay^{i_\lend - i_\lstart - i},~ b=0.93,
\end{equation}
where $\delta_{a,b}^{a',b'} := 1$ for  $(a, b)=(a', b')$ and $0$ otherwise; $C_p$ is the event generation threshold for polarity $p \in \{-1, +1\}$ and $i_\mathrm{start}$, $i_\mathrm{end}$ indicate the indices of the first and last events in $(t_0, t_1]$, respectively. The term $\decay^{i_\lend - i_\lstart - i}$, with $\decay := 0.93$ makes events that are far in the past weigh less than events that are rather recent. As we show in \cref{sec:decay} of our appendix, this is necessary to make our approach robust against noise events.

In \cref{eq:eventloss}, we defined $\mathcal{L}_\mathrm{event}$, which uses $E(t_0, t_1)$ for random $t_0$ and $t_1$ each training iteration.
As event streams could contain millions of events per second, implementing the accumulation procedure directly by processing each event on every query would be too slow to run every training iteration.
To overcome this issue, we use an approach inspired by Lin \etal~\cite{fastedi} to accelerate the queries:
We treat all pixels independently and store all intermediate accumulation results in arrays $A^{x_i,y_i}_{j}=E(0, t_i)_{x_i,y_i}$, and $T^{x_i,y_i}_{j}=t_i$, where $j$ is the per-pixel event counter.
Then, once we need to compute $E(t_0, t_1)$, we split it into two queries:
\begin{equation}
    E(t_0, t_1)=E(0, t_1)-E(0, t_0).
\end{equation}

To compute $E_{x,y}(0, t)$, we find the largest $j_{x,y}$ using binary search such that $T^{x_i,y_i}_{j}<t$, for each pixel $x,y$.
Then $E_{x,y}(0, t)=A^{x,y}_{j_{x,y}}$,
which reduces the temporal complexity of the query from $O(N_\mathrm{events})$ to $O(\log(N_\mathrm{events}))$.
In our tests, the improved accumulation only takes $\qtyrange{1}{2}{\milli\s}$ for all six views combined, while naive implementation requires $\qtyrange{4}{8}{\s}$ to compute the same queries.

\begin{figure}
    \centering

    \centering

   \bgroup
   \def\arraystretch{0.75}
   \setlength\tabcolsep{0.010\linewidth}

   \begin{tabularx}{0.95\columnwidth}{@{}>{\centering\arraybackslash}X>{\centering\arraybackslash}X>{\centering\arraybackslash}X>{\centering\arraybackslash}X>{\centering\arraybackslash}X@{}}
    \scriptsize \textbf{Ground truth} & \scriptsize \textbf{Our method} & \scriptsize E+Dyn-NeRF & \scriptsize B+Dyn-NeRF \\
    \includegraphics[width=\linewidth]{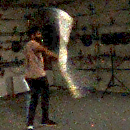} & \includegraphics[width=\linewidth]{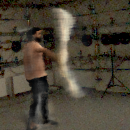} & \includegraphics[width=\linewidth]{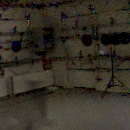} & \includegraphics[width=\linewidth]{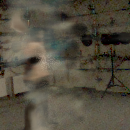}\\
    
    \includegraphics[width=\linewidth]{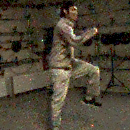} & \includegraphics[width=\linewidth]{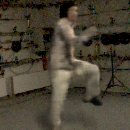} & \includegraphics[width=\linewidth]{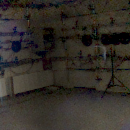} & \includegraphics[width=\linewidth]{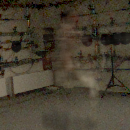} \\
    \scriptsize \textbf{Ground truth}&\scriptsize E+FreeNeRF~\cite{yang2023freenerf} & \scriptsize D+Dyn-NeRF & \scriptsize D+NR-NeRF\cite{tretschk2021nonrigid}\\
    \includegraphics[width=\linewidth]{fig/fig4_cvpr_crop1/207/gt/0260.png}&\includegraphics[width=\linewidth]{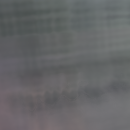} & \includegraphics[width=\linewidth]{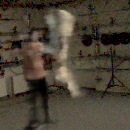} & \includegraphics[width=\linewidth]{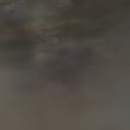}\\
    
    \includegraphics[width=\linewidth]{fig/fig4_cvpr_crop1/80/gt/0255.png}&\includegraphics[width=\linewidth]{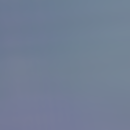} & \includegraphics[width=\linewidth]{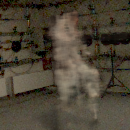} & \includegraphics[width=\linewidth]{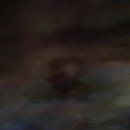}

\end{tabularx}
\egroup
\vspace{-0.4cm}

    \caption{%
    For two real scenes, we compare novel views by different methods (top: \enquote{Towel T.}, bottom: \enquote{Dancing}). \enquote{E+}, \enquote{B+}, and \enquote{D+} are trained using events processed with E2VID~\cite{Rebecq19pami}, using Blurry RGB frames, and using EDI-deblurred frames~\cite{pan2019bringing}, respectively. Dyn-NeRF represents our method supervised only with RGB.
    E2VID-based baselines fail due to view inconsistencies of E2VID data and its artefacts.
    Blurry RGB+Dyn-NeRF fails due to the blurriness and sparsity of its inputs.
    NR-NeRF fails due to the fast and large motion in the scenes.
    }
    \label{fig:realcomp}

\end{figure}

\begin{figure}
\centering
\bgroup
\def\arraystretch{0.75}
\setlength\tabcolsep{0.012\linewidth}
\begin{tabularx}{\columnwidth}{@{}>{\centering\arraybackslash}X >{\centering\arraybackslash}X >{\centering\arraybackslash}X@{}}
    \scriptsize \textbf{Ground truth} & \scriptsize \textbf{Our method} & \scriptsize E+Dyn-NeRF \\
    \includegraphics[width=\linewidth,trim={0 0 0 95pt},clip]{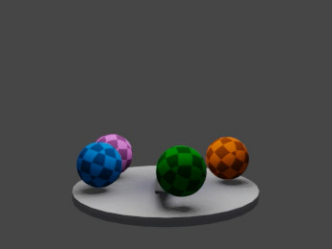}& \includegraphics[width=\linewidth,trim={0 0 0 95pt},clip]{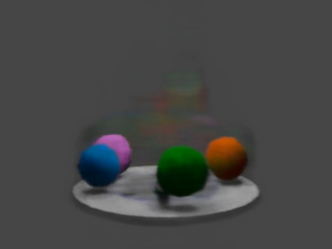}& \includegraphics[width=\linewidth,trim={0 0 0 95pt},clip]{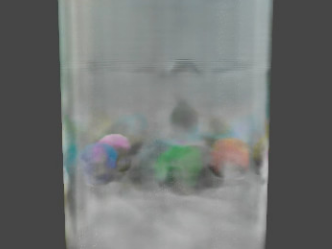}\\
    \scriptsize B+Dyn-NeRF & \scriptsize E+FreeNeRF~\cite{yang2023freenerf} & \scriptsize B+FreeNeRF~\cite{yang2023freenerf} \\
    \includegraphics[width=\linewidth,trim={0 0 0 95pt},clip]{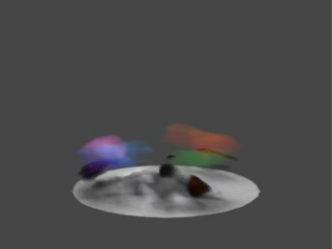} &
    \includegraphics[width=\linewidth,trim={0 0 0 95pt},clip]{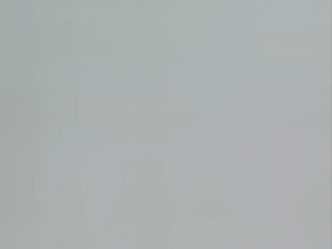} &
    \includegraphics[width=\linewidth,trim={0 0 0 95pt},clip]{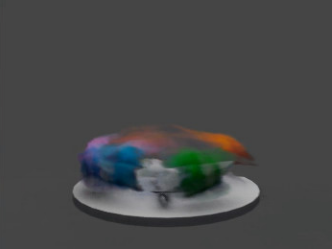}\\
    \scriptsize R+Dyn-NeRF & \scriptsize R+NR-NeRF~\cite{tretschk2021nonrigid} & \\
    \includegraphics[width=\linewidth,trim={0 0 0 95pt},clip]{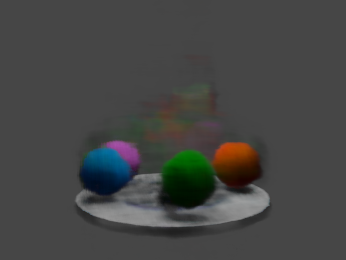}& \includegraphics[width=\linewidth,trim={0 0 0 95pt},clip]{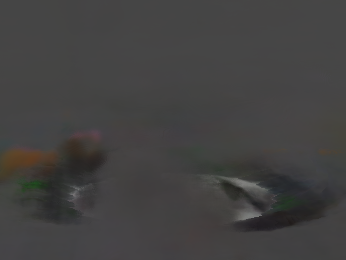}& 
\end{tabularx}
\egroup

\caption{Qualitative comparisons on synthetic ``Blender'' sequence. "E+", "B+", and "R+" baselines are trained with E2VID-processed events, blurry RGB frames, and original RGB frames as inputs, respectively. Dyn-NeRF is our method without event supervision using only RGB data. %
}
\label{fig:synthcomp}
\end{figure}

\section{Experiments}
We compare our method to RGB-based baselines that do not use events and are trained either on blurry RGB recordings or RGB videos generated from events using E2VID~\cite{Rebecq19pami}.
We evaluate on both synthetic and real sequences.
We evaluate the predicted novel views over the reconstructed sequence by computing PSNR, SSIM, and LPIPS.
For the synthetic data, we use three hold-out views to obtain quantitative results.

\subsection{Synthetic Dataset}
\label{sec:synthdataset}
To generate synthetic data, we render five scenes in Blender at $\qty{1000}{FPS}$ from five fixed views arranged uniformly around the object at the same height.
These are then fed into an event simulator \cite{rudnev2023eventnerf} to obtain event streams.
Additionally, we render all scenes using three more fixed views for use as the test set.
More details are in \cref{sec:datasetcomposition}. 

\subsection{Multi-view Event Camera Setup and Data}
\label{sec:realdataset}
To record the real dataset, we set up six hardware-synchronised iniVation DAVIS 346C cameras (\cref{fig:realsetup} in appx.).
We recorded $16$ sequences with different subjects, motions, and objects, totalling $\qty{18}{\min}$ of simultaneous multi-view RGB and event streams. 
Note that all  sequences were recorded at \num{7.0}\si{\lux}
to \num{11.5}\si{\lux} (from a lux-meter), \ie{} very dim, requiring $\qty{150}{\milli\s}$ exposure time for RGB frames to be of decent brightness, which in turn reduced the frame rate to $\qty{5}{FPS}$.  As a result, the RGB frames recorded by the DAVIS are very blurry (see \cref{fig:blurry} in the Appx.). 

We placed the cameras around the capture area at different heights, ranging from $\qtyrange[range-units=repeat, range-phrase=\text{ to }]{110}{250}{\centi\meter}$ centred on the subject.
Lens focal lengths were set to \qty{4}{\milli\m} to capture as much of the scene as possible.
The cameras were connected to a workstation using dedicated PCIe USB3 controller boards and fibre optic USB3 cables.
We modified the provided DV software~\cite{dvsoftware} to support saving 12-stream AEDAT~4 recordings (one RGB and one event stream for each of the six cameras).

To calibrate the cameras, we use a commercial multi-view motion capture software \cite{thecaptury} with reported 3D/2D reprojection errors under $\qty{2.2}{\milli\meter}$ in 3D and $\qty{0.2}{px}$ in 2D.
It requires synchronous RGB video streams of a moving chequerboard for calibration.
The event cameras can record both events and RGB frames simultaneously.
The shutter is activated automatically at the user-defined frame rate and exposure time, and precise trigger timestamps are saved with the frames.
However, the software does not allow for these frames to be triggered simultaneously.
To mitigate this, we recorded both RGB frames and events and synthesised synchronous multi-view RGB streams with Fast EDI~\cite{fastedi} debayered with VNG~\cite{vng}.
This allowed us to reconstruct deblurred RGB frames at the exact provided timestamps for each view.
\begin{figure}
    \centering
    \bgroup
    \def\arraystretch{0.75}
    \setlength\tabcolsep{0.006\linewidth}
    \begin{tabularx}{\columnwidth}{@{}>{\centering\arraybackslash}X>{\centering\arraybackslash}X>{\centering\arraybackslash}X>{\centering\arraybackslash}X>{\centering\arraybackslash}X@{}}
        \scriptsize Reference RGB & \scriptsize \textbf{Full Model} & \scriptsize TensoRF-CP~\cite{tensorf} & \scriptsize NGP~\cite{instantngp} & \scriptsize HexPlane~\cite{cao2023hexplane}\\
        \includegraphics[width=\linewidth]{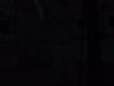} & \includegraphics[width=\linewidth]{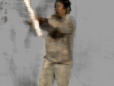} & \includegraphics[width=\linewidth]{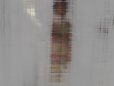} & \includegraphics[width=\linewidth]{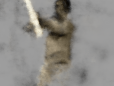} & \includegraphics[width=\linewidth]{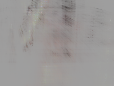} \\
        \scriptsize Enhanced RGB & \scriptsize w/o multi-seg. & \scriptsize w/o $L_\mathrm{sparsity}$ & \scriptsize w/o clipping & \scriptsize w/o decay \\
        
        \includegraphics[width=\linewidth]{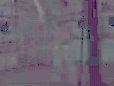} & \includegraphics[width=\linewidth]{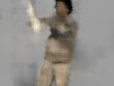} & \includegraphics[width=\linewidth]{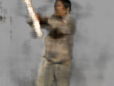} & \includegraphics[width=\linewidth]{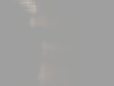} & \includegraphics[width=\linewidth]{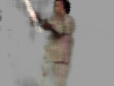} \\
        
        \scriptsize w/o $L_\mathrm{event}$ & \scriptsize w/o $L_\mathrm{acc}$ & \scriptsize w/o $L_\mathrm{RGB}$ & \scriptsize only $L_\mathrm{event}$ & \scriptsize only $L_\mathrm{acc}$\\
        \includegraphics[width=\linewidth]{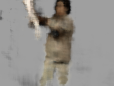} & \includegraphics[width=\linewidth]{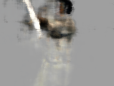} & \includegraphics[width=\linewidth]{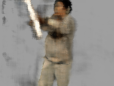} & \includegraphics[width=\linewidth]{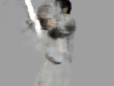} & \includegraphics[width=\linewidth]{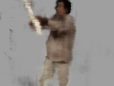} \\

    \end{tabularx}
    \egroup
    \caption{%
    Ablation study on the \enquote{Sword} real sequence.
    Our full method produces best quality predictions.
    Alternative backbones~\cite{tensorf,instantngp,cao2023hexplane} perform worse, as their grid-based structure does not allow for smooth propagation of details across time. Disabling various parts of the method reduces the prediction quality. 
    Especially disabling event decay makes the contours of the subject more blurry.
    The recordings were done in the dark, as apparent from a reference RGB video that was captured with a regular camera (not used in training). As it is pitch black, we also show the enhanced version.%
    }
    \label{fig:ablreal}

\end{figure}

\begin{figure}
    \centering
    \bgroup
    \def\arraystretch{0.75}
    \setlength\tabcolsep{0.006\linewidth}
    \begin{tabularx}{\columnwidth}{@{}>{\centering\arraybackslash}X>{\centering\arraybackslash}X>{\centering\arraybackslash}X>{\centering\arraybackslash}X@{}}
        \scriptsize Ground truth & \scriptsize TensoRF-CP~\cite{tensorf} & \scriptsize NGP~\cite{instantngp} & \scriptsize HexPlane~\cite{cao2023hexplane}\\
        \includegraphics[width=\linewidth]{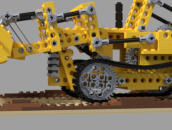} & 
        \includegraphics[width=\linewidth]{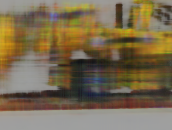} & 
        \includegraphics[width=\linewidth]{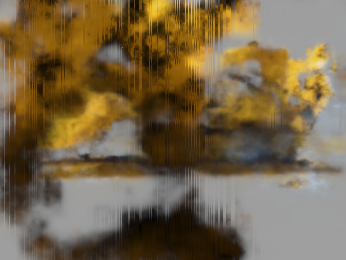} & 
        \includegraphics[width=\linewidth]{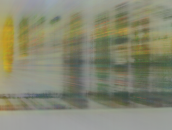} \\
        \scriptsize \textbf{Full Model} & \scriptsize w/o $L_\mathrm{event}$ & \scriptsize w/o $L_\mathrm{acc}$ & \scriptsize w/o $L_\mathrm{RGB}$ \\
        \includegraphics[width=\linewidth]{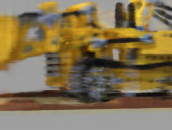} & 
        \includegraphics[width=\linewidth]{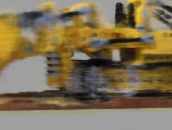} & 
        \includegraphics[width=\linewidth]{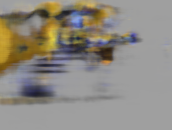} & 
        \includegraphics[width=\linewidth]{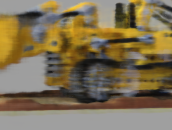} \\
        \scriptsize only $L_\mathrm{event}$ & \scriptsize only $L_\mathrm{acc}$ & \scriptsize w/o $L_\mathrm{sparsity}$ & \scriptsize w/o clipping \\
        \includegraphics[width=\linewidth]{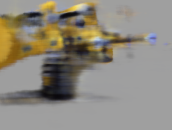} & 
        \includegraphics[width=\linewidth]{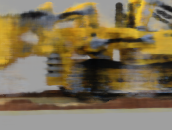} &
        \includegraphics[width=\linewidth]{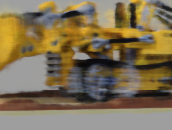} & 
        \includegraphics[width=\linewidth]{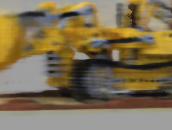}
    \end{tabularx}
    \egroup

    \caption{%
    Ablation studies on the ``Static Lego'' synthetic sequence (zoomed in). NGP fails to produce correct geometry in our sparse-view setting. TensoRF-CP and HexPlane fail due to their grid-based nature preventing propagation of details through time. Disabling parts of the method results in blurrier predictions and artefacts.
    }
    \label{fig:ablsynth}

\end{figure}

\begin{table*}
    \centering
    \sisetup{detect-family=true, text-series-to-math = true, propagate-math-font = true, round-mode=places,round-precision=2}
\resizebox{1.00\textwidth}{!}{
    \begin{tabular}{l|S|S|S|S|S|S|S|S|S|S|S|S|S|S|S|S|S|S}
 &
  \multicolumn{3}{c|}{Blender} &
  \multicolumn{3}{c|}{Dress} &
  \multicolumn{3}{c|}{Spheres} &
  \multicolumn{3}{c|}{Lego} &
  \multicolumn{3}{c|}{Static Lego} &
  \multicolumn{3}{c}{Average} \\
  \hline
 Method&
  PSNR $\uparrow$&
  SSIM $\uparrow$&
  LPIPS $\downarrow$&
  PSNR $\uparrow$&
  SSIM $\uparrow$&
  LPIPS $\downarrow$&
  PSNR $\uparrow$&
  SSIM $\uparrow$&
  LPIPS $\downarrow$&
  PSNR $\uparrow$&
  SSIM $\uparrow$&
  LPIPS $\downarrow$&
  PSNR $\uparrow$&
  SSIM $\uparrow$&
  LPIPS $\downarrow$&
  PSNR $\uparrow$&
  SSIM $\uparrow$&
  LPIPS $\downarrow$ \\
  \hline
Our Dyn-EventNeRF &
  {\bfseries$\num{27.60780654}$} &
  {\bfseries$\num{0.9298977988}$} &
  {\bfseries$\num{0.1125655338}$} &
  {\bfseries$\num{31.14645113}$} &
  {\bfseries$\num{0.950413243}$} &
  {\bfseries$\num{0.07675376749}$} &
  {\bfseries$\num{29.84148519}$} &
  {\bfseries$\num{0.9241641747}$} &
  {\bfseries$\num{0.08266463444}$} &
  {\bfseries$\num{22.41513639}$} &
  {\bfseries$\num{0.799447964}$} &
  {\bfseries$\num{0.2858156326}$} &
  {\bfseries$\num{23.94315044}$} &
  {\bfseries$\num{0.844104826}$} &
  0.1966507311 &
  {\bfseries$\num{26.99080594}$} &
  {\bfseries$\num{0.8896056013}$} &
  {\bfseries$\num{0.1508900599}$} \\
{E2VID~\cite{Rebecq19pami} + Dyn-NeRF} &
  12.4732891 &
  0.675842645 &
  0.5364007041 &
  21.90022458 &
  0.8797760933 &
  0.3325069591 &
  19.10660288 &
  0.8706128369 &
  0.2718079382 &
  18.20098549 &
  0.724539447 &
  0.4519101699 &
  15.77211655 &
  0.6971556703 &
  0.5979456007 &
  17.49064372 &
  0.7695853385 &
  0.4381142744 \\
{Blurry RGB + Dyn-NeRF} &
  23.14883185 &
  0.8890725093 &
  0.1514781723 &
  27.41289499 &
  0.9159519361 &
  0.1373602989 &
  20.53894801 &
  0.8779331122 &
  0.2114722642 &
  20.2858448 &
  0.7493470848 &
  0.4038061529 &
  22.99980619 &
  0.8192695181 &
  0.2072387519 &
  22.87726517 &
  0.8503148321 &
  0.2222711281 \\
{E2VID~\cite{Rebecq19pami} + FreeNeRF~\cite{yang2023freenerf}} &
  9.069228188 &
  0.6269393922 &
  0.4990051697 &
  19.20822043 &
  0.8921230777 &
  0.2860587507 &
  17.56010201 &
  0.8747407029 &
  0.2834805152 &
  17.83739561 &
  0.7269743935 &
  0.3815842296 &
  15.6938898 &
  0.7200885587 &
  0.5639533207 &
  15.87376721 &
  0.768173225 &
  0.4028163972 \\
{Blurry RGB + FreeNeRF~\cite{yang2023freenerf}} &
  24.87724255 &
  0.9064988428 &
  0.1188490811 &
  29.65766389 &
  0.9418883665 &
  0.07958486051 &
  19.74971482 &
  0.87447393 &
  0.222323164 &
  20.05650558 &
  0.7552789549 &
  0.3198580266 &
  22.27759451 &
  0.8292697468 &
  {\bfseries$\num{0.1672369935}$} &
  23.32374427 &
  0.8614819682 &
  0.1815704252\\
{GT RGB~\cite{pan2019bringing} + Dyn-NeRF} &
  25.683 &
  0.912 &
  0.159 &
  30.234 &
  0.944 &
  0.106 &
  28.132 &
  0.915 &
  0.107 &
  22.088 &
  0.794 &
  0.312 &
  22.840 &
  0.819 &
  0.228 &
  25.795 &
  0.877 &
  0.182 \\ 
{GT RGB~\cite{pan2019bringing} + NR-NeRF~\cite{tretschk2021nonrigid}} &
  19.98760946061006 &
  0.836801776720316 &
  0.3940494442143456 &
  28.789877356542238 &
  0.9343943567395776 &
  0.14914222164909438 &
  16.415376865165 &
  0.84245624486412 &
  0.377088615406464 &
  14.742602446641108 &
  0.6855590393898441 &
  0.602584305081037 &
  14.527485245416347 &
  0.6607096620897244 &
  0.5809638956592421 &
  18.8925902748749 &
  0.791984215960716 &
  0.420765696402037 \\  
  \multicolumn{15}{c}{}
\end{tabular}
}
\resizebox{1.0\textwidth}{!}{
\begin{tabular}{l|S|S|S|S|S|S|S|S|S|S|S|S|S|S|S|S}
&
  \multicolumn{4}{c|}{Dancing (real)} &
  \multicolumn{4}{c|}{Bucket (real)} &
  \multicolumn{4}{c|}{Towel Tricks (real)} &
  \multicolumn{4}{c}{Average (real)} \\
  \hline
 Method&
  ROI~PSNR $\uparrow$&
  PSNR $\uparrow$&
  SSIM $\uparrow$&
  LPIPS $\downarrow$&
  ROI~PSNR $\uparrow$&
  PSNR $\uparrow$&
  SSIM $\uparrow$&
  LPIPS $\downarrow$&
  ROI~PSNR $\uparrow$&
  PSNR $\uparrow$&
  SSIM $\uparrow$&
  LPIPS $\downarrow$&
  ROI~PSNR $\uparrow$&
  PSNR $\uparrow$&
  SSIM $\uparrow$&
  LPIPS $\downarrow$ \\
  \hline
Our Dyn-EventNeRF &
\bfseries 22.25 &
\bfseries 28.5751583698805 &
\bfseries 0.814792092848126 &
\bfseries 0.136793960183859 &
\bfseries 24.45 &
\bfseries 29.5623907010004 &
\bfseries 0.825172266504358 &
\bfseries 0.117926983237267 &
\bfseries 20.82 &
\bfseries 27.5977156470529 &
\bfseries 0.8083705186589 &
\bfseries 0.135122140571475 &
\bfseries 22.51 &
\bfseries 28.5784215726446 &
\bfseries 0.816111626003795 &
\bfseries 0.1299476946642 \\
{E2VID~\cite{Rebecq19pami} + Dyn-NeRF} &
 15.64 &
  21.2009013777531 &
0.704781823646836 &
0.387421740293503 &
 16.42 &
18.3978052673332 &
0.661475154382704 &
0.451675924360752 &
 16.71 &
22.7803036426688 &
0.732067838703574 &
0.307592697441578 &
 16.26 &
20.7930034292517 &
0.699441605577705 &
0.382230120698611 \\
{Blurry RGB + Dyn-NeRF} &
 18.88 &
  26.0214262433698 &
0.787472315151304 &
0.181774728000164 &
 21.58 &
28.0847698591356 &
0.810105238127948 &
0.135202926620841 &
 15.42 &
22.0598350728621 &
0.735972632006754 &
0.246245318502188 &
 18.63 &
25.3886770584558 &
0.777850061762002 &
0.187740991041064 \\
{E2VID~\cite{Rebecq19pami} + FreeNeRF~\cite{yang2023freenerf}} &
 10.99 &
  9.34566294735015 &
0.234185938471091 &
1.06199187040329 &
 9.72 &
8.99735377007147 &
0.221940116537096 &
1.03368803716841 &
 14.67 &
13.0936961529529 &
0.259799485800892 &
1.0951146364212 &
 11.79 &
10.4789042901248 &
0.23864184693636 &
1.06359818133097\\
{EDI~\cite{pan2019bringing} + Dyn-NeRF} &
 19.407062619026355 &
  26.73422620544163 &
0.79592480718628 &
0.15641547657549382 &
 21.85138738976197 &
28.31204884243635 &
0.8114410917226574 &
0.13517406590282918 &
 18.01260794100201 &
25.532457205697288 &
0.7887930354829773 &
0.15452582463622094 &
 19.7570193165968 &
26.8595774178584 &
0.798719644797305 &
0.148705122371515\\
{EDI~\cite{pan2019bringing} + NR-NeRF~\cite{tretschk2021nonrigid}} &
 14.986377826056088 &
  15.487755160777871 &
0.24175634708911697 &
0.95651725679636 &
 16.598653829189093 &
15.100904103838802 &
0.295238913563557 &
1.0407376459666662 &
 15.353822247858819 &
15.816823033775522 &
0.23923388755895297 &
1.0483032315969467 &
 15.646284634368 &
15.4684940994641 &
0.258743049403876 &
1.01518604478666
\end{tabular}
}
 
    \caption{We compare our method to the baselines on both the synthetic (top) and real (bottom) datasets by rendering novel views and computing scores against the reference images for those views. Dyn-NeRF refers to the proposed Dyn-EventNeRF method with only RGB supervision. Our method produces higher output quality than other methods, consistently across all scenes. ROI scores were computed using a tight per-frame foreground mask estimated from the ground truth. GT RGB is using ground-truth non-blurry training views as input. Despite that, NR-NeRF cannot produce accurate results as the motion is too fast and large for it to learn deformations.
    Zoom recommended. 
    }
    \label{tbl:hybrid}
\end{table*}

\begin{table}
    \centering

    \sisetup{detect-family=true, text-series-to-math = true, propagate-math-font = true, round-mode=places,round-precision=2}
    \resizebox{7.0cm}{!}{
\begin{tabular}{l|S[round-mode=places,round-precision=3]|S[round-mode=places,round-precision=3]|S[round-mode=places,round-precision=3]}
 Method & PSNR $\uparrow$ & SSIM $\uparrow$ & LPIPS $\downarrow$ \\
    \hline
Ours (TensoRF-CP~\cite{tensorf}) & 24.167333584754587 & 0.8391663366331918 & 0.28531962737441063 \\
Ours (NGP~\cite{instantngp}) & 23.23942254441942 & 0.7765649873852443 & 0.32740966933468973 \\
Ours (HexPlane~\cite{cao2023hexplane}) & 21.69418947682925 & 0.803876806325189 & 0.3638652837773164 \\
\hline
w/o clipping & 24.470577445092772 & 0.864222437643462 & 0.24007930236558117 \\
w/o $L_\mathrm{event}$ & 24.51052901025229 & 0.8638161678926994 & 0.2358033755173286 \\
w/o $L_\mathrm{acc}$ & 25.11305080305545 & 0.8598775386685107 & 0.20414959562321505 \\
w/o $L_\mathrm{RGB}$ & 26.593070219085888 & 0.8856899655677692 & 0.15888269870231547 \\
only $L_\mathrm{event}$ & 25.525104707466888 & 0.8640451176734286 & 0.20215818845977385 \\
only $L_\mathrm{acc}$ & 25.79536539317553 & 0.876945422335534 & 0.18243062607944013 \\
w/o $L_\mathrm{sparsity}$ & 26.85316814010856 & 0.886839200463549 & 0.15708070765559873 \\
\hline
Our Full Model (Final) & \bfseries 27.09671166078875 & \bfseries 0.893037778571142 & \bfseries 0.1454134922660887 \\
    \end{tabular}
    }
     \caption{Quantitative ablation and design choice study averaged over all synthetic scenes. All metrics clearly favour our full model.
    }
    \label{tbl:ablation}
\end{table}

\subsection{Implementation}
\label{ssec:implementation}
We base our code on EventNeRF~\cite{rudnev2023eventnerf}.
One model takes $\num{e5}$ iterations and $\num{6}$ hours to converge on a single NVIDIA A40 GPU.
Hence, a sequence of $10$ models takes $60$ GPU-hours of training, and we use $10$ GPUs for our experiments.
Note that comparable RGB-based approaches \cite{park2021nerfies,park2021hypernerf} require much, much more training time (32--1344 GPU hours).
For more details, refer to \cref{sec:mlpdetail,sec:baselinedetail}.

\subsection{Comparisons}\label{sec:comparison}
We compare our method with three RGB-based baselines: Our Dynamic EventNeRF with pure RGB supervision (abbr.: \mbox{Dyn-NeRF}), deformation-based NR-NeRF~\cite{tretschk2021nonrigid}, single-frame sparse-view FreeNeRF~\cite{yang2023freenerf} trained per each frame of input data.
These baselines are applied to the blurry RGB streams and RGB frames reconstructed from events using E2VID~\cite{Rebecq19pami}. Additionally, we run the methods on exact rendered RGB input with synthetic data, and frames deblurred with EDI~\cite{pan2019bringing} with real data.
For synthetic sequences, following ~\cite{rudnev2023eventnerf}, we measure PSNR, SSIM, and LPIPS for novel views throughout the sequence.
We evaluate the predictions at $\qty{20}{FPS}$, and render $3$ fixed hold-out views for each of these moments.
For real data, we use $3$ sequences from our dataset and choose one fixed view as the ground truth, leaving the other $5$ as the training views.
Since the subject only takes a small portion of the frame in some real sequences, we also report masked PSNR within the tight foreground mask (ROI PSNR), computed using the ground-truth background captures.

Our Dynamic EventNeRF (abbr.: Dyn-EventNeRF)
outperforms all of the baselines (\cref{tbl:hybrid}). 
\cref{fig:synthcomp,fig:realcomp} confirm this visually: %
All baselines based on blurry RGBs fail to recover sharp details as they were designed to work with sharp data.
In contrast, our proposed Dyn-EventNeRF method uses events instead, which allows it to recover sharp details.
Methods trained on RGB frames generated from events using E2VID and EDI also recover these details.
However, similarly to Ref.~\cite{rudnev2023eventnerf}, there are many artefacts in the reconstructions, as all input views were reconstructed independently, disregarding multi-view consistency.
Our method integrates multi-view events into one shared volume, allowing for view-consistent 3D reconstruction and reduced artefacts.
NR-NeRF~\cite{tretschk2021nonrigid} fails to reconstruct most of the scenes. Due to the extreme amplitude and speed of the motion and view sparsity in the tested sequences, it can learn neither the deformations nor the canonical volume correctly.
\subsection{Ablation and Design Choice Study}
\label{ssec:ablations}
We conduct an ablation and design choice study (\cref{tbl:ablation,fig:ablsynth,fig:ablreal}; per-scene in \cref{tbl:ablationfull} in the appendix) using both synthetic and real data.
In particular, we compare different options for the core model: MLP (\enquote{Full Model}), NGP~\cite{instantngp}, \mbox{TensoRF-CP}~\cite{tensorf}, and \mbox{HexPlane}~\cite{cao2023hexplane}.
NGP fails to handle the sparse-view setting.
TensoRF-CP and HexPlane are both grid-based: Individual timestamps are modelled separately, so that information from one timestamp cannot propagate to other timestamps, leading to blurriness and artefacts.
On the other hand, our full model with temporally-conditioned MLP explicitly controls how much information is shared between timestamps through positional encoding.
Supervising one timestamp also affects others in its vicinity, resulting in better use of the training data.
Disabling $\mathcal{L}_\text{event}$, $\mathcal{L}_\text{acc}$, or $\mathcal{L}_\text{RGB}$ leads to increased blurriness because disabling either of them will reduce the amount of supervision:
Without $\mathcal{L}_\text{event}$, there is no way to correct relative changes between predictions through events, without $\mathcal{L}_\text{acc}$ there is no way to propagate sharp RGB information and static parts of the scene. %
And without $\mathcal{L}_\text{RGB}$, the model is not encouraged to predict the reference RGB frames correctly.
Disabling both $\mathcal{L}_\text{RGB}$ and $\mathcal{L}_\text{acc}$ removes the entire RGB supervision.
As the model only uses the events, there can be an exposure ambiguity, leading to lower metrics.
Disabling both $\mathcal{L}_\text{RGB}$ and $\mathcal{L}_\text{event}$ leaves only $\mathcal{L}_\text{acc}$.
Despite using both RGB and events, this baseline performs worse than the full model, presumably due to accumulation artefacts.
Cylinder clipping prevents the model from creating geometry in regions that are only supervised by one view.
If disabled, the model produces numerous artefacts in the novel views; in the case of real data specifically, it even fails to converge.
$\mathcal{L}_\text{sparsity}$ forces the model to reduce the number of opaque pixels in the reconstruction.
Removing it leads to semi-transparent floating artefacts in the reconstructed volume.
Using only one single model for the entire sequence duration (as opposed to dividing the sequence into shorter chunks) results in a blurrier reconstruction.
In this case, artefacts produced by reconstructing one of the segments can propagate to other segments, which is not the case in the multi-segment model.

\section{Conclusion}

We introduced Dynamic EventNeRF---the first approach to 4D reconstruction of dynamic, non-rigid scenes from multi-view event streams augmented with sparse RGB frames. 
Our approach outperforms all tested baselines, producing state-of-the-art reconstructions of fast motions under dim and dark lighting conditions. 
Our ablation study shows that our full model, comprising events, RGB, sparsity, event accumulation losses and cylinder clipping performs best. In particular, we found the MLP backbone to be the most suitable for the task. 

In a further qualitative ablation study (see \cref{sec:viewcnt}), we have analysed the impact of the number of viewpoints on the reconstruction quality. The experiment indicates that quality is indeed increased by adding additional views.
This can serve as a justification for further research into multi-view event camera setups for 4D reconstruction.

Adding external high-resolution RGB cameras to the event setup could further improve the sharpness of the rendered novel views.
Future work could also investigate architectures that achieve similar quality at a smaller cost (\eg~a variant of 3DGS~\cite{kerbl20233dgs}).

We assembled and calibrated a multi-view event camera setup to record, to the best of our knowledge, the first dataset that consists of six-view simultaneous events and RGB recordings of complex motions in challenging lighting conditions.
We made this dataset available online, such that despite the limited availability of multi-view event camera data, further research can continue to explore this modality.

{
    \small
    \bibliographystyle{ieeenat_fullname}
    \bibliography{main}
}

\clearpage
\appendix
\maketitlesupplementary

\begin{figure}[h]
    \centering
    \includegraphics[width=1.0\columnwidth]{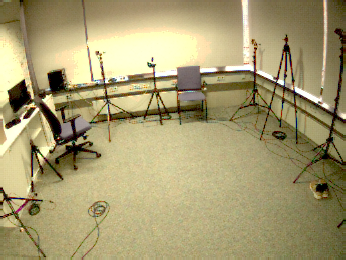}
    \caption{Our portable setup in one of the recording rooms. It consists of six hardware-synchronised iniVation DAVIS 346C colour event cameras on tripods, connected to a workstation with \qty{10}{\m} optic fibre USB3 extension cables. We installed two additional PCIe USB3 extension cards into the workstation to connect all cameras with the required bandwidth.}
    \label{fig:realsetup}
\end{figure}
\begin{figure}[t]
    \centering
    \includegraphics[width=1.0\linewidth,angle=180,origin=c]{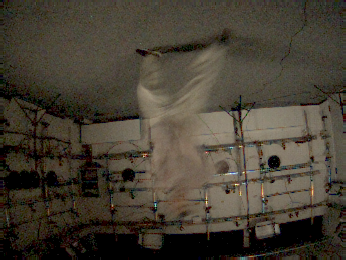}
    \caption{RGB frames recorded by the DAVIS camera are very blurry, because low-light conditions require longer exposure times.}
    \label{fig:blurry}
\end{figure}

This appendix provides additional experiments and details on calibration, hyperparameters and baselines. 
We show how we calibrate our event camera response function in \cref{sec:crf}.
In \cref{sec:mlpdetail}, we describe the architecture of our MLP network.
Next, \cref{sec:baselinedetail} includes additional details on the baselines and their hyperparameters.
We then describe our datasets and explain the capture process in \cref{sec:datasetcomposition}.
We demonstrate the performance of our method with additional ablations on the number of supporting RGB frames (\cref{sec:suprgb}), number of views (\cref{sec:viewcnt}), and provide the full per-scene ablation results in \cref{tbl:ablationfull}. Finally, we discuss event accumulation stability and the proposed accumulation decay, proving it effective and correct in our case (\cref{sec:decay}).

\begin{figure}[t]
\begin{subfigure}{\linewidth}
    \centering
    \includegraphics[width=1.0\linewidth]{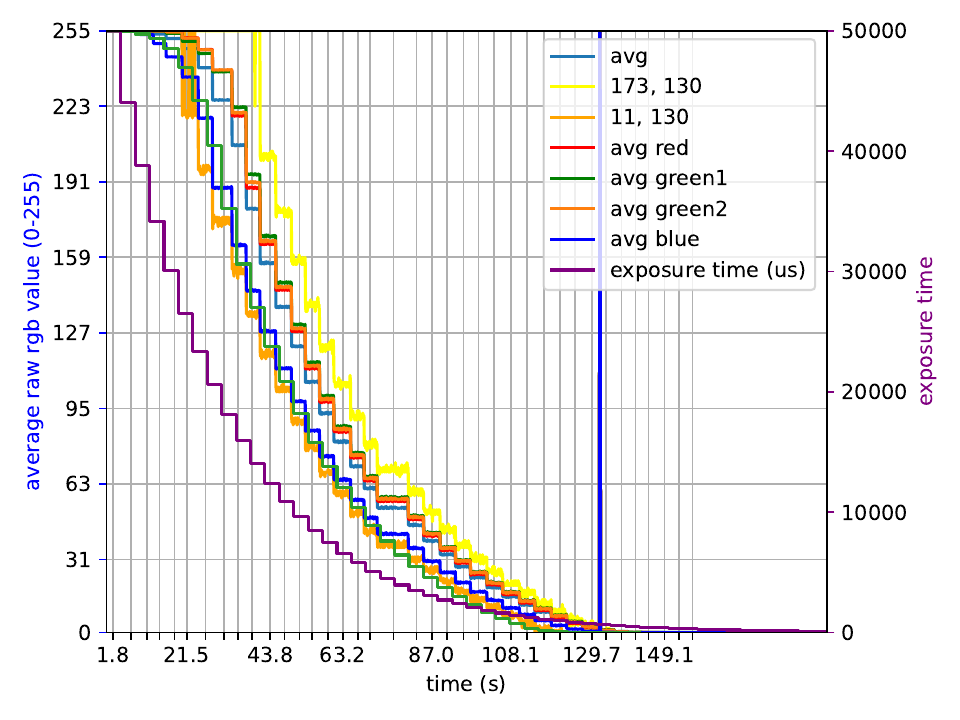}
    \caption{Raw recorded RGB values when varying the exposure time of the camera at different pixel locations and averaged over the colour channels. There is an outlier on the right of the blue channel average curve, which we ignore during calibration.}
    \label{fig:crf1}
\end{subfigure}
\hfill
\begin{subfigure}{\linewidth}
    \centering
    \includegraphics[width=1.0\linewidth]{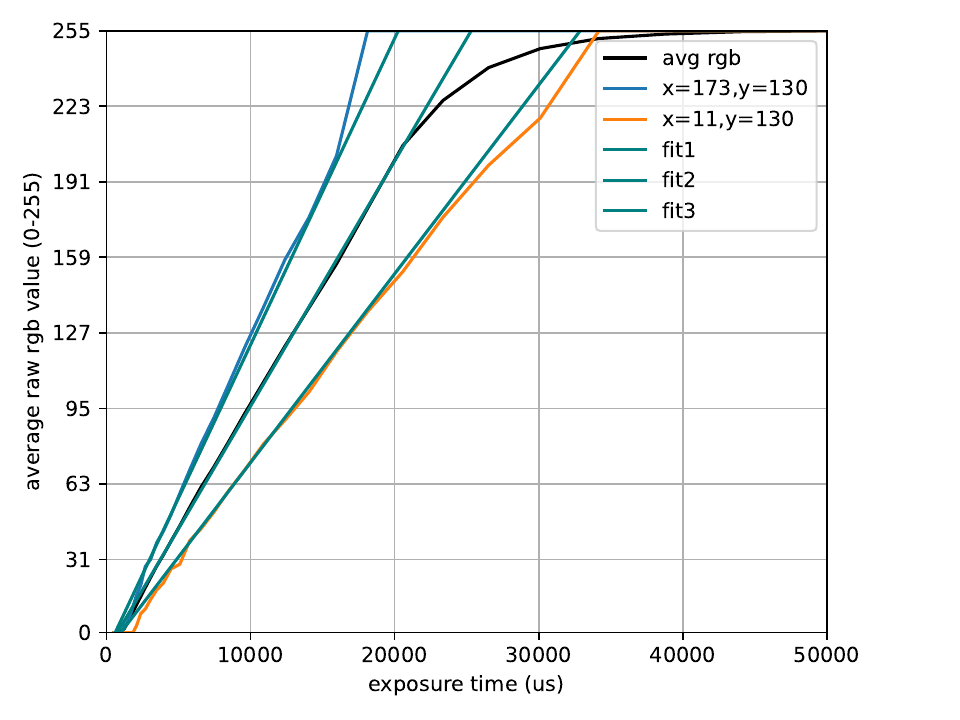}
    \caption{Measured CRF and our linear fit. \enquote{avg rgb} is the RGB value averaged over all pixels in the view. Lens vignetting results in the values close to white (255) being rolled off. To mitigate this issue, we use RGB values of single pixels instead, labelled as \enquote{$x=\hdots, y=\hdots$} on the plot. \enquote{fit1}, \enquote{fit2}, and \enquote{fit3} indicate our respective linear fits to these curves with $\epsilon=\num{3e-2}$ shift over the y axis (in 0-1 range; corresponds to $7.65$ in 0-255 range of the plot).}
    \label{fig:crf2}
\end{subfigure}
\caption{Event camera RGB intensity frame CRF calibration.}
\label{fig:crf}
\end{figure}

\section{RGB Frame Quality}

As a result of the low-light conditions (see \cref{sec:realdataset}), the exposure durations of the RGB frames recorded by the DAVIS camera are rather long, resulting in severe motion blur (see \cref{fig:blurry}).

\section{Camera CRF Calibration}
\label{sec:crf}
To combine the event generation model with the RGB intensity frames, both obtained through the same lens with  DAVIS~346C event cameras, we need a precise measurement of the camera response function (CRF) that we obtain as follows:
First, we place the camera in front of a constant brightness light source.
Then we use the fact that the amount of captured light is directly proportional to the exposure time.
The DAVIS~346 software allows varying the exposure time at $\si{\micro\s}$ precision.
Thus, by varying it, we can record the relative amount of light needed for the recorded pixel intensity to reach any value from $0$ to $255$.

We show the raw measurements in \cref{fig:crf1}.
The CRF is obtained by plotting the RGB values over the exposure, which we show in \cref{fig:crf2}.
Due to the vignetting of the lens and view-dependent effects, different pixel locations respond differently.
Because of that, averaging the values from different pixel locations results in smooth roll-offs at the extremes of the RGB values, which do not correspond to the actual sensor properties.
The results show that the measured CRF can be approximated well as a linear function with a vertical shift of $\epsilon=\num{3e-2}$ over the y-axis, indicating that RGB value $\num{0}$ can be reported even when a non-zero amount of light reaches the sensor.

\begin{table}[t]
    \centering

    \sisetup{detect-family=true, text-series-to-math = true, propagate-math-font = true, round-mode=places,round-precision=2}
    \resizebox{7.0cm}{!}{
\begin{tabular}{l|S[round-mode=places,round-precision=3]|S[round-mode=places,round-precision=3]|S[round-mode=places,round-precision=3]}
 Method & PSNR $\uparrow$ & SSIM $\uparrow$ & LPIPS $\downarrow$ \\
    \hline
Ours (TensoRF-CP~\cite{tensorf}) & 26.20190745 & 0.784033381 & 0.163783091 \\
Ours (NGP~\cite{instantngp}) & 26.27821244 & 0.787437711 & 0.163224690 \\
Ours (HexPlane~\cite{cao2023hexplane}) & 24.94064475 & 0.758267750 & 0.209537897 \\
\hline
w/o clipping & 25.28794029 & 0.809762055 & 0.162667827 \\
w/o decay & 27.11897488 & 0.815725305 & 0.133418845 \\
w/o multi-segment & 26.67297292 & 0.809257575 & 0.142403389 \\
w/o $L_\mathrm{sparsity}$ & 26.91965485 & 0.813790626 & 0.127648863 \\
w/o $L_\mathrm{event}$ & 27.61980746 & \bfseries 0.820358338 & 0.122953489 \\
w/o $L_\mathrm{acc}$ & 25.51727716 & 0.801984987 & 0.139127600 \\
w/o $L_\mathrm{RGB}$ & 27.06160713 & 0.818148033 & \bfseries 0.120109750 \\
only $L_\mathrm{event}$ & 25.02855487 & 0.798512370 & 0.142937280 \\
only $L_\mathrm{acc}$ & \bfseries 27.75413693 & \bfseries 0.821195091 & 0.121601953 \\
only $L_\mathrm{RGB}$ & 26.019370863660694 & 0.8059220532150707 & 0.14449443958699704 \\
\hline
Our Full Model (Final) & 27.04764140 & \bfseries 0.818727174 & \bfseries 0.119746349 \\
    \end{tabular}
    }
     \caption{Quantitative ablation and design choice study on the ``Sword'' real scene. SSIM and LPIPS metrics favour our full model. In particular, omitting event decay ("w/o decay") makes SSIM and LPIPS scores worse.
    }
    \label{tbl:ablation_real}
\end{table}

\begin{table*}[ht]
    \centering

    \sisetup{detect-family=true, text-series-to-math = true, propagate-math-font = true, round-mode=places,round-precision=3}
    \resizebox{0.95\textwidth}{!}{
    \begin{tabular}{l|S|S|S|S|S|S|S|S|S}
                                                 & \multicolumn{3}{c|}{Blender}                                                                                     & \multicolumn{3}{c|}{Dress}                                                                                       & \multicolumn{3}{c}{Spheres}                                                                                      \\
\hline
Method                                           & PSNR $\uparrow$                     & SSIM $\uparrow$                     & LPIPS $\downarrow$                   & PSNR $\uparrow$                     & SSIM $\uparrow$                     & LPIPS $\downarrow$                   & PSNR $\uparrow$                    & SSIM $\uparrow$                     & LPIPS $\downarrow$                    \\
\hline
TensoRF-CP~\cite{tensorf}       & 24.726813089903864                  & 0.8794343719216627                  & 0.22677601898709934                  & 28.09059424681952                   & 0.9149561143295492                  & 0.20216893777251244                  & 23.971144235363905                 & 0.8662842272266924                  & 0.27954707369208337                   \\
NGP~\cite{instantngp}           & 25.687043317672543                  & 0.8883473644725631                  & 0.18415225570400556                  & 29.131450056670456                  & 0.9333993586254541                  & 0.17891080640256404                  & 28.75452290138639                  & 0.9146406662325446                  & 0.12018560022115707                   \\
HexPlane~\cite{cao2023hexplane} & 23.119187666592207                  & 0.8636399331966244                  & 0.2747482453783353                   & 23.580262647709404                  & 0.8664528464244375                  & 0.33556743984421095                  & 21.636131887597458                 & 0.8531302130713831                  & 0.33395645916461947                   \\
\hline
w/o clipping                                     & 19.9585240027306                    & 0.8512668433748166                  & 0.4259934296210607                   & 27.926075186536856                  & 0.9257616562470338                  & 0.14659005819509427                  & 28.785698739645664                 & 0.9162527041530875                  & 0.09319814884414275                   \\
w/o $L_\mathrm{sparsity}$                        & \bfseries 27.500890010806586 & 0.9264032506500594                  & 0.13455955541382233                  & 31.292436937216575                  & 0.9500758114377508                  & 0.08149800902853409                  & 29.357744778639542                 & 0.92018954856145                    & 0.08555545300866167                   \\
w/o $L_\mathrm{event}$                           & 19.9585240027306                    & 0.8512668433748166                  & 0.4259934296210607                   & 29.883649189577003                  & 0.9414686416951644                  & 0.10806965318818887                  & 27.744175527732093                 & 0.9131522926810878                  & 0.11381753807266554                   \\
w/o $L_\mathrm{acc}$                             & 26.834836707336915                  & 0.8994623856548549                  & 0.12490875832736492                  & 30.64954977131498                   & 0.9464376801596031                  & 0.0788711549093326                   & 28.846532259455543                 & 0.9198019246096365                  & 0.08703924504419168                   \\
w/o $L_\mathrm{RGB}$                             & 27.32070335986759                   & 0.927934694661158                   & 0.12191798351705074                  & 30.540144287499256                  & 0.9457583545012865                  & 0.08565744670728842                  & 29.128532219304418                 & 0.9203454192245266                  & 0.08892725923409064                   \\
only $L_\mathrm{event}$        & 27.361913593041248                  & 0.9053265133636219                  & 0.14791633263230325                  & 31.3077732422317                    & 0.9506241389846629                  & \bfseries 0.0727877560382088  & 29.44676905192386                  & 0.9206393182086023                  & 0.08358714915812016                   \\
only $L_\mathrm{acc}$     & 25.683241879154767                  & 0.9122951063253973                  & 0.158982390537858                    & 30.234220903611504                  & 0.9439512053821882                  & 0.10572639523694913                  & 28.13223267245972                  & 0.915299381578368                   & 0.1073509798074762                    \\
\hline
Full Model                                       & 27.430956492478717                  & \bfseries 0.9290692165945287 & \bfseries 0.11914152161528667 & \bfseries 31.395668613445324 & \bfseries 0.9515512222015783 & 0.07590672944982847                  & \bfseries 29.76533383493896 & \bfseries 0.9241795851040315 & \bfseries 0.08055415963754058  \\
\multicolumn{10}{c}{}                                                                                                                                                                                                                                                                                                                                                                                     \\
                                                 & \multicolumn{3}{c|}{Lego}                                                                                        & \multicolumn{3}{c|}{Static Lego}                                                                                 & \multicolumn{3}{c}{Average}                                                                                      \\
\hline
Method                                           & PSNR $\uparrow$                     & SSIM $\uparrow$                     & LPIPS $\downarrow$                   & PSNR $\uparrow$                     & SSIM $\uparrow$                     & LPIPS $\downarrow$                   & PSNR $\uparrow$                    & SSIM $\uparrow$                     & LPIPS $\downarrow$                    \\
\hline
TensoRF-CP~\cite{tensorf}       & 21.444033233458914                  & 0.7516677647088329                  & 0.40797935575246813                  & 22.604083118226722                  & 0.7834892049792221                  & 0.3101267506678899                   & 24.167333584754587                 & 0.8391663366331918                  & 0.28531962737441063                   \\
NGP~\cite{instantngp}           & 17.134322523579634                  & 0.594455333820939                   & 0.5562706371148427                   & 15.48977392278808                   & 0.5519822137747208                  & 0.5975290472308795                   & 23.23942254441942                  & 0.7765649873852443                  & 0.32740966933468973                   \\
HexPlane~\cite{cao2023hexplane} & 20.09318391195744                   & 0.7110994518527959                  & 0.4519244134426117                   & 20.042181270289745                  & 0.7250615870807041                  & 0.42312986105680467                  & 21.69418947682925                  & 0.803876806325189                   & 0.3638652837773164                    \\
\hline
w/o clipping                                     & 22.99582224461544                   & 0.8135067044857712                  & 0.27247931286692617                  & 22.68676705193529                   & 0.8143242799566006                  & 0.2621355623006821                   & 24.470577445092772                 & 0.864222437643462                   & 0.24007930236558117                   \\
w/o $L_\mathrm{sparsity}$                        & 22.416481922556216                  & 0.7984959340242205                  & 0.28113740831613543                  & 23.698287051323874                  & 0.8390314576442641                  & 0.2026531125108401                   & 26.85316814010856                  & 0.886839200463549                   & 0.15708070765559873                   \\
w/o $L_\mathrm{event}$                           & 21.97989425370418                   & 0.7919762091770549                  & 0.30545998190840085                  & 22.986402077517603                  & 0.8212168525353742                  & 0.22567627479632696                  & 24.51052901025229                  & 0.8638161678926994                  & 0.2358033755173286                    \\
w/o $L_\mathrm{acc}$                             & \bfseries 23.17775427285463  & \bfseries 0.8200897386185336 & \bfseries 0.2467940221230189  & 16.0565810043152                    & 0.7135959642999252                  & 0.4831347977121671                   & 25.11305080305545                  & 0.8598775386685107                  & 0.20414959562321505                   \\
w/o $L_\mathrm{RGB}$                             & 22.33156284364386                   & 0.7985870151644703                  & 0.29210372095306714                  & 23.644408385114314                  & 0.8358243442874042                  & 0.2058070831000805                   & 26.593070219085888                 & 0.8856899655677692                  & 0.15888269870231547                   \\
only $L_\mathrm{event}$        & 23.009271055062577                  & 0.8144922235833785                  & 0.2623805207510789                   & 16.499796595075036                  & 0.7291433942268778                  & 0.44411918371915815                  & 25.525104707466888                 & 0.8640451176734286                  & 0.20215818845977385                   \\
only $L_\mathrm{acc}$      & 22.087511700456236                  & 0.7937204003514341                  & 0.31189005002379416                  & 22.839619810195437                  & 0.8194610180402816                  & 0.22820331479112307                  & 25.79536539317553                  & 0.876945422335534                   & 0.18243062607944013                   \\
\hline
Full Model                                       & 23.029062230350398                  & 0.8180886648971114                  & 0.2583080090582371                   & \bfseries 23.862537132730342&\bfseries 0.8423002040584598&\bfseries0.19315704156955082&\bfseries27.09671166078875&\bfseries0.893037778571142&\bfseries0.1454134922660887
\end{tabular}
}
     \caption{Quantitative ablation and design choice study done with all synthetic scenes. While some ablated models performed better in single scenes, the averaged metrics clearly favour our full model.
    }
    \label{tbl:ablationfull}
\end{table*}

\section{MLP Architecture and Hyperparameters}
\label{sec:mlpdetail}
We inherit the NeRF MLP architecture used in EventNeRF~\cite{rudnev2023eventnerf}, which we also optimise with Adam~\cite{2015KingmaBa}.
However, we modify the model to use the same shared network for both fine and coarse levels of prediction rather than using separate ones.
This allows for better optimisation stability and speed, as only half as many parameters are optimised, compared to the original version.
Due to the small number of input views in our setting, we modified the code to compose training batches such that they contain an equal number of rays from each view.
We found this to increase the stability of the training and the accuracy of the predictions.

\section{Baseline Implementation Details}
\label{sec:baselinedetail}
NGP, TensoRF-CP and HexPlane were reimplemented on top of our codebase.
For NGP, we used a hash grid implementation in tiny-cuda-nn~\cite{tiny-cuda-nn}.
For the hash-grid encoding, we used the following configuration:
\begin{verbatim}
    "otype": "HashGrid",
    "n_levels": 8,
    "n_features_per_level": 2,
    "log2_hashmap_size": 19,
    "base_resolution": 8,
    "per_level_scale": 2.0
\end{verbatim}
For the subsequent MLP network, we use two layers with 16 hidden and 10 geometry features.
Then, for the colour network, we use three layers with 64 hidden features.
In total, we train the NGP method for $\num{3d4}$ iterations.
The resulting model diverged when training in the sparse-view setting.
To significantly improve its sparse-view performance, we annealed the cylinder bound radius from 0 to $100\%$ of the full value in the first \num{d4} iterations.
Despite that, its sparse-view performance is still lacking compared to the full model and other ablated architectures.
TensoRF-CP was reimplemented from scratch in PyTorch.
In addition to the usual three spatial dimensions, we also decomposed the temporal dimension, turning it into a spatio-temporal representation.
We started with a $16\times 16\times 16\times 16$ grid and gradually progressed towards a $500\times 500 \times 500 \times 24$ grid in 10 steps throughout $\num{2d3}$ iterations.
We set the factorisation rank to $8$ as the highest value that did not cause out-of-memory errors with our NVIDIA A40 GPU.
In total, we train the method for $\num{d4}$ iterations.
Similarly, HexPlane was also reimplemented from scratch.
We also used a $500\times 500\times 500\times 24$ grid with a factorisation rank of $8$.

\section{Dataset Composition}
\label{sec:datasetcomposition}
The proposed synthetic dataset consists of
\begin{enumerate}
    \item Three new original scenes: ``Spheres'', ``Blender'', ``Dress'' (licensed CC-BY4.0), and
    \item Two scenes that were based on the data provided in \cite{Mildenhall19}: ``Lego'', ``Static Lego''.
\end{enumerate}

The proposed real dataset contains over $\qty{18}{\minute}$ of simultaneous multi-view event and RGB frame streams, recorded on our six event-camera rig described in Sec.~\ref{sec:realdataset}.

We captured ten subjects.
Each of them was instructed about the recording and signed the release form for the use of the recorded data in our experiments and subsequent public release.
The instructions were as follows: \enquote{Enter the recording area. Run around the centre of the area, perform kicks, punches, jumping jacks, and then whatever fast motions you like for a total of a minute. Afterwards, take one of the available objects (towel, ball, bass guitar, paper poster \enquote{sword}, box, bucket) and perform fast motions for about another minute.}

\begin{figure*}[h]
    \centering
   \bgroup
   \def\arraystretch{0.75}
   \setlength\tabcolsep{0.004\linewidth}
\begin{tabularx}{\textwidth}{>{\centering\arraybackslash}X >{\centering\arraybackslash}X >{\centering\arraybackslash}X >{\centering\arraybackslash}X >{\centering\arraybackslash}X >{\centering\arraybackslash}X
>{\centering\arraybackslash}X}
    0.5 FPS & 1 FPS & \textbf{5 FPS} & 10 FPS & 50 FPS & 100 FPS & Ground Truth \\
    \includegraphics[width=\linewidth]{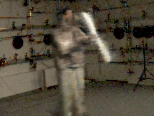}&
    \includegraphics[width=\linewidth]{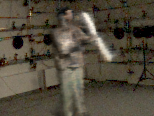}&
    \includegraphics[width=\linewidth]{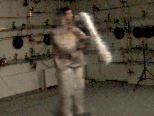}&
    \includegraphics[width=\linewidth]{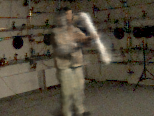}&
    \includegraphics[width=\linewidth]{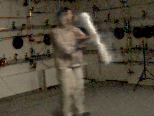}&
    \includegraphics[width=\linewidth]{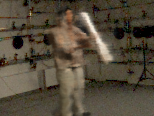}&
    \includegraphics[width=\linewidth]{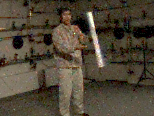}\\

    \includegraphics[width=\linewidth]{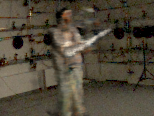}&
    \includegraphics[width=\linewidth]{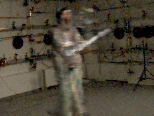}&
    \includegraphics[width=\linewidth]{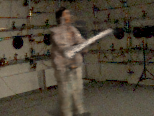}&
    \includegraphics[width=\linewidth]{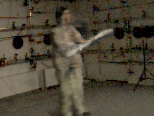}&
    \includegraphics[width=\linewidth]{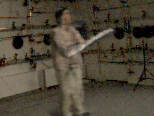}&
    \includegraphics[width=\linewidth]{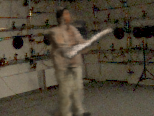}&
    \includegraphics[width=\linewidth]{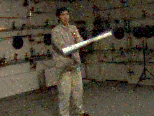}
\end{tabularx}
    \egroup
\caption{Ablation on the number of supporting RGB images used for training. We show novel views from two different times in two rows; bold indicates the default value. The 0.5 FPS model uses only one set of RGB images. As the number of images increases, there is a slight reduction in artefacts. However, even with one set of images (0.5 FPS), the results are close to the full model (5 FPS). This indicates that the method uses primarily event information and does not rely on the RGB images much.}
\label{fig:ablfps}
\end{figure*}

\begin{figure*}[h]
    \centering
   \bgroup
   \def\arraystretch{0.75}
   \setlength\tabcolsep{0.004\linewidth}
\begin{tabularx}{\textwidth}{>{\centering\arraybackslash}X >{\centering\arraybackslash}X >{\centering\arraybackslash}X >{\centering\arraybackslash}X >{\centering\arraybackslash}X}
    2 views & 3 views & 4 views & \textbf{5 views} & Ground Truth \\
    \includegraphics[width=\linewidth]{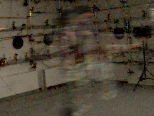} &
    \includegraphics[width=\linewidth]{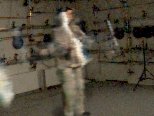} &
    \includegraphics[width=\linewidth]{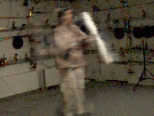} &
    \includegraphics[width=\linewidth]{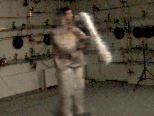} &
    \includegraphics[width=\linewidth]{fig/aaa/0224/gt.png} \\

    \includegraphics[width=\linewidth]{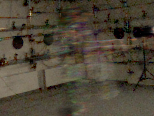} &
    \includegraphics[width=\linewidth]{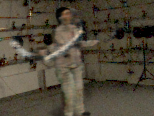} &
    \includegraphics[width=\linewidth]{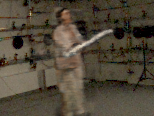} &
    \includegraphics[width=\linewidth]{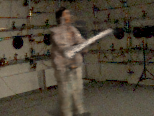} &
    \includegraphics[width=\linewidth]{fig/aaa/0270/0270_gt.png}
\end{tabularx}
    \egroup
\caption{Ablation on the number of views used for training. We show novel views at two different times in two rows; bold indicates the default value. With more input views, the quality indeed significantly improves.} 
\label{fig:ablviews}
\end{figure*}
\begin{figure*}[t]
    \centering
   \bgroup
   \def\arraystretch{0.75}
   \setlength\tabcolsep{0.004\linewidth}
\begin{tabularx}{0.85\textwidth}{>{\centering\arraybackslash}X >{\centering\arraybackslash}X >{\centering\arraybackslash}X >{\centering\arraybackslash}X}
    2 views & 3 views & 4 views & 6 views \\
    \includegraphics[width=\linewidth]{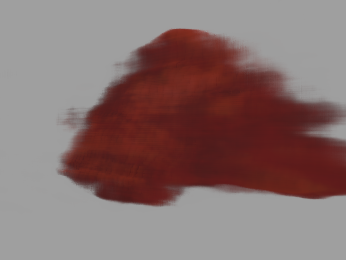} &
    \includegraphics[width=\linewidth]{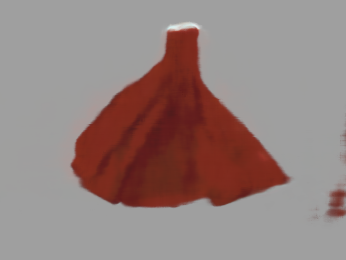} &
    \includegraphics[width=\linewidth]{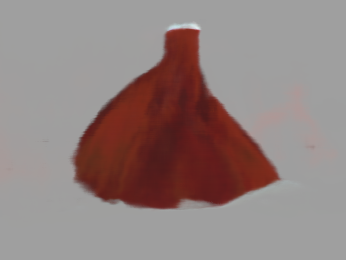} &
    \includegraphics[width=\linewidth]{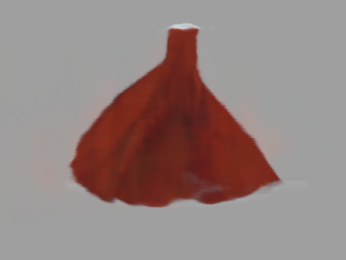} \\
    12 views & 24 views & 36 views & Ground Truth\\
    \includegraphics[width=\linewidth]{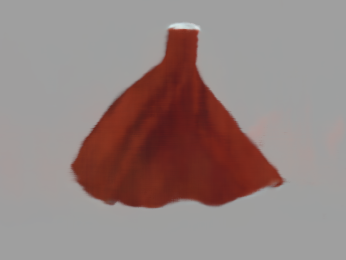} &
    \includegraphics[width=\linewidth]{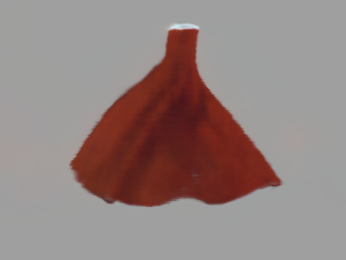} &
    \includegraphics[width=\linewidth]{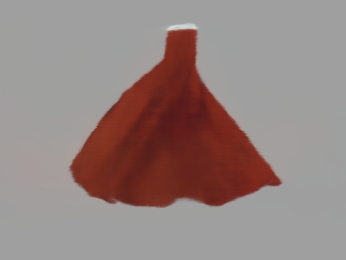} &
    \includegraphics[width=\linewidth]{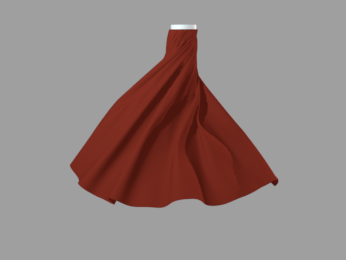}
\end{tabularx}
    \egroup
\caption{Additional synthetic-data ablation on the number of views used for training. With more input views, the visual quality improves, albeit hitting diminishing returns at 24 views.} 
\label{fig:ablviews_synth}
\end{figure*}

\section{Ablation on FPS of Supporting RGB Frames}
\label{sec:suprgb}

We ablate the number of supporting RGB frames used for training (\cref{fig:ablfps,tbl:ablation_fps}).
There is only a minimal difference in the results if we use only one RGB frame for reconstruction (0.5 FPS), compared to using 100 FPS RGB inputs.
This indicates that our method does not depend on the RGB inputs much, using mostly the events.
That could lead to a follow-up work that eliminates the RGB inputs.

\section{Ablations on View Count}
\label{sec:viewcnt}

When reducing the number of training views, we see that the model does not diverge even when using only three views (\cref{fig:ablviews,tbl:ablation_view}).
However, we note that the accuracy of geometry increases significantly when using more views.
We also used synthetic data to test more possible setups, up to 36 views.
There is a clear improvement in PSNR with the increase in the number of training views as well.
This confirms that our method indeed benefits from additional improvements to the setup, as stated in the conclusion.

\section{Ablation on Design Choices}
We provide a detailed ablation study of our design choices in \cref{tbl:ablationfull}, listing quantitative results for all our synthetic scenes individually. \cref{tbl:ablationfull} clearly shows that our full model performs best overall.
We also provide a similar table for one of the real scenes in \cref{tbl:ablation_real}.
SSIM and LPIPS also clearly favour the full model in that case.

\begin{table}
    \centering

    \sisetup{detect-family=true, text-series-to-math = true, propagate-math-font = true, round-mode=places,round-precision=2}
    \resizebox{6.0cm}{!}{
\begin{tabular}{l|S[round-mode=places,round-precision=3]|S[round-mode=places,round-precision=3]|S[round-mode=places,round-precision=3]}
 FPS & PSNR $\uparrow$ & SSIM $\uparrow$ & LPIPS $\downarrow$ \\
    \hline
0.5 & 26.950          & 0.817          & 0.123            \\
1   & 26.932          & 0.817          & 0.123           \\
2   & 27.055          & 0.818          & 0.119          \\
5   & 27.061          & \bfseries 0.820 & \bfseries 0.115  \\
10  & 27.289          & \bfseries 0.821 & 0.117           \\
20  & 27.226          & \bfseries 0.821 & \bfseries 0.114   \\
30  & \bfseries 27.337 & \bfseries 0.821 & 0.116            \\
50  & 27.310          & \bfseries 0.820 & 0.117          \\
100 & \bfseries 27.328 & \bfseries 0.821 & 0.116  
    \end{tabular}
    }
     \caption{Quantitative study on the number of supporting RGB frames. By default, we use 5 FPS, the same frequency as our raw data. SSIM and LPIPS metrics favour values above or equal to 5 FPS, but do not drop too much with fewer frames, indicating that the model primarily uses event information and not RGB.
    }
    \label{tbl:ablation_fps}
\end{table}

\begin{table}[h]
    \centering

    \sisetup{detect-family=true, text-series-to-math = true, propagate-math-font = true, round-mode=places,round-precision=2}
    \resizebox{7.0cm}{!}{
\begin{tabular}{l|S[round-mode=places,round-precision=3]|S[round-mode=places,round-precision=3]|S[round-mode=places,round-precision=3]}
 Input Views & PSNR $\uparrow$ & SSIM $\uparrow$ & LPIPS $\downarrow$ \\
    \hline
2     & 24.667          & 0.784          & 0.166          \\
3     & 26.081          & 0.808          & 0.149          \\
4     & 26.891          & 0.815          & 0.118          \\
5     & \bfseries 27.061 & \bfseries 0.820 & \bfseries 0.115
    \end{tabular}
    }
     \caption{Quantitative study on the number of input views on the ``Sword'' real scene. All metrics clearly confirm the improvement with additional views.
    }
    \label{tbl:ablation_view}
\end{table}

\begin{table}[h]
    \centering

    \sisetup{detect-family=true, text-series-to-math = true, propagate-math-font = true, round-mode=places,round-precision=2}
    \resizebox{\linewidth}{!}{
    \begin{tabular}{c|c|c|c|c|c|c}
        2 v. & 3 v. & 4 v. & 6 v. & 12 v. & \textbf{24 v.} & \textbf{36 v.}\\
        \hline
        \num{16.05}&\num{28.97}&\num{27.90}&\num{30.88}&\num{32.19}&{\bfseries \num{33.36}}&{\bfseries \num{33.27}}\\
    \end{tabular}
    }
     \caption{PSNR scores on \enquote{Dress} with varying number of views. As stated in the conclusion, our method could easily benefit from an increase in the number of views or the resolution of the cameras, which is evident from the results. Bold indicates the two best-performing model results.
    }
    \label{tbl:ablation_view_synth}
\end{table}

\section{Accumulation Stability}
\label{sec:decay}
\begin{figure}
    \centering
    \includegraphics[width=0.595\columnwidth,trim={0.5cm 0.5cm 0.5cm 0.5cm},clip]{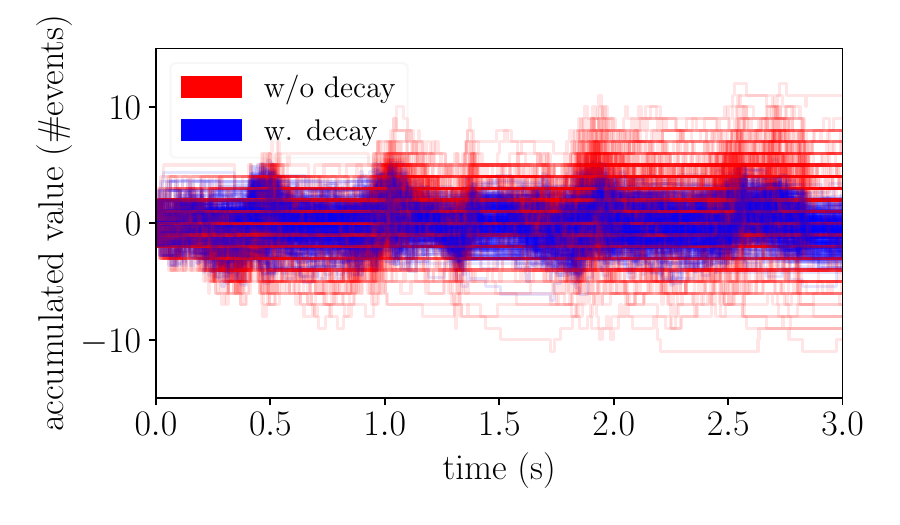}
    \includegraphics[width=0.385\columnwidth,trim={0.2cm 0.2cm 0.2cm 0.2cm},clip]{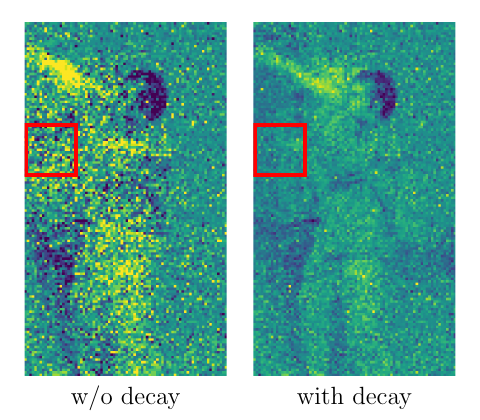}
    \caption{%
        Demonstration of accumulation decay (\cref{ssec:accumulation}). In a small patch of pixels (marked red on the right), we accumulate events for each pixel individually and show the resulting signals. Naive method (red) becomes unstable, as all pixels have completely different values at the end. In contrast, our proposed method with decay (blue) is stable: all pixels keep similar values at any time. Visually, the image with decay (right) has much fewer artefacts, too.
    }
    \label{fig:damping}

\end{figure}

We have observed that accumulating long sequences of events leads to unstable results. This means that even in a pixel of constant brightness, spurious \enquote{noise} events will accumulate to ever-increasing deviations from the true brightness level.
In this section, we prove the existence of this phenomenon analytically. We also show that our decay approach successfully limits this problem to a tolerable bound.

To see that noise events can destroy all information about the true brightness level, let us consider a pixel of constant brightness that, nevertheless, reports noise events with polarities $\polarity_i \in\{-1, + 1\}$. For the sake of simplicity, we assume that the $\polarity_i$ are independent and identically distributed random variables with fixed, but arbitrary probabilities $\pprobpos := \probability(\polarity_i = +1)$ and $\pprobneg := \probability(\polarity_i = -1)$.
The accumulated polarity after $\numevents$ such events can then be expressed as the random variable 
\begin{equation}
\accpol_\numevents := \sum\limits^\numevents_{i = 1} \polarity_i, 
\end{equation}
which is a simplified version of \cref{eq:accumulation}. The expected value of $\accpol_\numevents$ is
\begin{equation}
    \expectation(\accpol_\numevents) \jeq{\text{Def.} \expectation} \sum\limits^\numevents_{i = 1} (+1) \cdot \pprobpos + (-1) \cdot \pprobneg = \numevents \cdot (\pprobpos - \pprobneg).
\end{equation}

If $\pprobpos \neq \pprobneg$, then $\lim\limits_{\numevents \rightarrow \infty} \expectation(\accpol_\numevents) = \pm \infty$, \ie{} as one accumulates more and more noise, one can safely expect to drift arbitrarily far away from the true brightness level. But even when $\pprobpos = \pprobneg$, which would make $\expectation(\accpol_\numevents) = 0$ for all $\numevents$, the \emph{variance} of (and therefore the confidence in) the accumulated polarity will decrease with growing $\numevents$:

\begin{equation}\label{eq:drowning}
    \begin{array}{lcl}
    &\variance(\accpol_\numevents) 
    
     \jeq{\text{Def.} \variance} &\expectation((\accpol_\numevents - \expectation(\accpol_\numevents))^2) \\
    
    &\contract\jeq{\text{Binom. thm.}}&\contract  \expectation (\accpol_\numevents^2 - 2 \accpol_\numevents \cdot \expectation(\accpol_\numevents) + \expectation(\accpol_\numevents)^2) \\
    
    &\contract\jeq{\text{Additivity} \; \expectation}&\contract  \expectation (\accpol_\numevents^2) - \expectation(\accpol_\numevents)^2 \\
    
    &\contract\jeq{\text{Def.} \accpol_\numevents}&\contract  \expectation \left(\sum\limits_{\substack{1 \leq i \leq \numevents  \\ 1 \leq j \leq \numevents }} \polarity_i \polarity_j \right) - \expectation(\accpol_\numevents)^2 \\
    
    &\contract\jeq{\text{Arithmetic}}&\contract  \expectation \left(\sum\limits_{i = 1} \polarity_i^2 + 2 \cdot\sum\limits_{1 \leq i < j \leq \numevents  } \polarity_i \polarity_j\right) - \expectation(\accpol_\numevents)^2 \\

    &\contract\jeq{\text{Additivity} \; \expectation}&\contract   \sum\limits_{i = 1} \expectation(\polarity_i^2) + 2 \cdot\sum\limits_{1 \leq i < j \leq \numevents } \expectation(\polarity_i \polarity_j) - \expectation(\accpol_\numevents)^2 \\

    &\contract\jeq{\text{Def.} \expectation}& \contract \numevents  \cdot (\pprobpos + \pprobneg) \\
     && \contract + 2 \cdot\sum\limits_{1 \leq i < j \leq \numevents } \pprobpos^2 - 2 \pprobpos \pprobneg + \pprobneg^2  \\
     && \contract - \numevents  \cdot (\pprobpos - \pprobneg)\\

    &\contract\jeq{\substack{\text{Binom. thm.} \\ \text{Arithmetic}}}&\contract  2 (\pprobpos - \pprobneg)^2 \cdot\sum\limits_{i = 1}^{\numevents -1} i + 2 \numevents  \pprobneg\\

    &\contract\jeq{\substack{\text{Gauß sum} \\ \text{Arithmetic}}}&\contract \numevents \cdot (\numevents - 1) \cdot (\pprobpos - \pprobneg)^2  + 2 \numevents \pprobneg.
    
\end{array}
\end{equation}

This term can exceed any arbitrary bound if one lets $\numevents$ grow large enough:
If $\pprobneg = 0$, then $\pprobpos = 1$ and $\variance(\accpol_\numevents) = \numevents^2 - \numevents$. 
If $\pprobpos - \pprobneg = 0$, then $\pprobpos = \pprobneg = \frac{1}{2}$ and $\variance(\accpol_\numevents) = \numevents$. 
In all other cases we have $\pprobneg > 0$ and $(\pprobpos - \pprobneg)^2 > 0$ and thus also $\lim\limits_{\numevents \rightarrow \infty} \variance(\accpol_\numevents) = + \infty$.

This shows that the noise events will drown out all information about the actual brightness level, if one just waits long enough. Since this effect would deteriorate not only background pixels (which should not cause any events), but also foreground pixels (where noise is interleaved with legitimate events), we need to achieve $\lim\limits_{\numevents \rightarrow \infty} \variance(\accpol_\numevents) \in \reals$
and thus introduce decay:
When accumulating polarities, we value older events less than younger events:
\begin{equation}
\hat{\accpol}_\numevents := \sum\limits^\numevents_{i = 1} \polarity_i \cdot \decay^{\numevents - i},
\end{equation}
where the decay strength $\decay := 0.93$ was empirically found to be a useful value.  
Now we have:
\begin{equation}
    \begin{array}{lcl}
    \expectation(\hat{\accpol}_\numevents) 
    
    &\jeq{\text{Additivity} \; \expectation}& \sum\limits_{i=1}^\numevents \expectation(\polarity_i) \cdot \decay^{n - i} \\

    &\jeq{\text{Def.} \; \expectation}& (\pprobpos - \pprobneg) \cdot \sum\limits_{i=1}^\numevents  \decay^{\numevents - i} \\
    
    &\jeq{\text{Arithmetic}}& (\pprobpos - \pprobneg) \cdot  \sum\limits_{i=0}^{\numevents -1} \decay^i \\

    &\jeq{\text{Geometric sum}}& (\pprobpos - \pprobneg) \cdot  \frac{\decay^\numevents - 1}{\decay - 1}.
    
\end{array}
\end{equation}
Since $\lim\limits_{\numevents \rightarrow \infty} \decay^\numevents = 0$, we have that $\lim\limits_{\numevents \rightarrow \infty} \expectation(\hat{\accpol}_\numevents) $ is finite. 
In a derivation similar to \cref{eq:drowning}, we obtain

\begin{equation}\label{eq:variance_damp}
    \begin{array}{lcl}
    \variance(\hat{\accpol}_\numevents) 
    
     &\jeq{\substack{\text{Def.} \variance \\ \text{Binom. thm.} \\ \text{Additivity} \expectation}} &  \expectation (\hat{\accpol}_\numevents^2) - \expectation(\hat{\accpol}_\numevents)^2 \\
    
     &\jeq{\substack{\text{Def.} \; \hat{\accpol}_\numevents \\ \text{Arithmetic} \\ \text{Additivity} \expectation}} &  \sum\limits_{i = 1}^\numevents \expectation(\polarity_i^2) \cdot \decay^{2(\numevents - i)} \\
     && + 2 \cdot\sum\limits_{1 \leq i < j \leq \numevents } \expectation(\polarity_i \polarity_j) \cdot \decay^{\numevents - i} \cdot \decay^{\numevents - j} \\
     && - \expectation(\hat{\accpol}_\numevents)^2 \\

     &\jeq{\substack{\text{Def.} \expectation \\ \text{Binom. thm.} \\ \text{Arithmetic}}} & (\pprobpos + \pprobneg) \cdot   \sum\limits_{i = 0}^{\numevents - 1} (\decay^2)^i \\
     && + 2 \cdot (\pprobpos - \pprobneg)^2 \cdot \sum\limits_{i = 0}^{\numevents  -1}\decay^i \cdot  \sum\limits_{j = 0}^{i - 1}\decay^j \\
     && - \expectation(\hat{\accpol}_\numevents)^2 \\

     &\jeq{\text{Geometric sum}} & (\pprobpos + \pprobneg) \cdot   \frac{\decay^{2\numevents} - 1}{\decay^2 - 1} \\
     && + 2 \cdot (\pprobpos - \pprobneg)^2 \cdot \sum\limits_{i = 0}^{\numevents  -1}\decay^i \cdot  \frac{\decay^i - 1}{\decay - 1} \\
     && - \expectation(\hat{\accpol}_\numevents)^2 \\

     &\jeq{\text{Arithmetic}} & (\pprobpos + \pprobneg) \cdot   \frac{\decay^{2\numevents} - 1}{\decay^2 - 1} \\
     && + 2 \cdot \frac{(\pprobpos - \pprobneg)^2}{\decay - 1} \cdot \left(\sum\limits_{i = 0}^{\numevents  -1}(\decay^2)^i -  \sum\limits_{i = 0}^{\numevents  -1}\decay^i\right) \\
     && - \expectation(\hat{\accpol}_\numevents)^2 \\

     &\jeq{\text{Geometric sum}} & (\pprobpos + \pprobneg) \cdot  \frac{\decay^{2\numevents} - 1}{\decay^2 - 1} \\
     && + 2 \cdot \frac{(\pprobpos - \pprobneg)^2}{\decay - 1} \cdot (\frac{\decay^{2\numevents} - 1}{\decay^2 - 1} -  \frac{\decay^\numevents - 1}{\decay - 1} ) \\
     && - \expectation(\hat{\accpol}_\numevents)^2.

\end{array}
\end{equation}
All summands in the last term of \cref{eq:variance_damp} converge to a finite number as $\numevents$ grows larger, so  $\lim\limits_{\numevents \rightarrow \infty} \variance(\hat{\accpol}_\numevents) \in \reals$. This shows that decay is effectively bounding the deviation from the true brightness level to a finite error, even for arbitrary numbers of noise events. 

\cref{fig:damping} (right) shows that after applying decay, accumulated events become clear, even though they were unrecognisable without decay, proving the effectiveness of our strategy. Additionally, we take a small patch of that view and plot the accumulated values with and without decay in \cref{fig:damping} (left): We see that, as predicted analytically, without decay, the accumulation grows beyond all bounds due to the noise, while with decay, it remains in a constant range, exactly as the signal should be.

The smaller one chooses $\decay$, the smaller 
$\lim\limits_{\numevents \rightarrow \infty} \variance(\hat{\accpol}_\numevents)$ becomes and the faster it converges. However, since decay affects not only noise events but also legitimate ones,  we have to explain why it does not distort the actual signal too much. To see this, consider a single pixel: As long as the foreground is not moving through this pixel, it shows a constant level of background brightness. All events it emits are noise, and can safely be suppressed by decay (because the accumulated polarity is supposed to be zero). When the foreground enters the pixel, it will trigger a number of legitimate events. As long as these are recent, the decay will not suppress them too much, so $\hat{\accpol}$ is approximately at the foreground brightness level.
Assuming that $\decay$ is chosen suitably, motion is usually fast enough that the object will leave the pixel again before decay \enquote{undoes} the entry events completely. Leaving the pixel again triggers a set of legitimate events that are first sufficiently recent not to be suppressed by damping. As they move further into the past, so do the entry events, so $\hat{\accpol}$ will then correctly approximate the background brightness level again. Of course, decay does still negatively impact the legitimate events (especially when motion is occasionally slower than what $\decay$ was tuned for, such that the foreground is inside the pixel for longer durations), but the underlying MLP, supervised by all our losses, can compensate for that sufficiently.
 The ablations in \cref{tbl:ablation_real,fig:ablreal} show that SSIM scores, LPIPS and visual results are indeed improved by our decay technique.

\end{document}